\documentclass{article}

\usepackage{microtype}
\usepackage{graphicx}
\usepackage{subfigure}
\usepackage{booktabs}
\usepackage{hyperref}
\usepackage{xcolor}

\usepackage[accepted]{template/icml2020}

\usepackage[english]{babel}
\usepackage[T1]{fontenc}
\usepackage[utf8]{inputenc}
\usepackage{amsmath}
\usepackage{amsfonts}
\usepackage{amssymb}
\usepackage{cases}
\usepackage{commath}
\usepackage{mathtools}
\usepackage{nicefrac}
\usepackage{stmaryrd}
\usepackage{bm}
\usepackage{etoolbox}
\usepackage{import}
\usepackage{transparent}
\usepackage{url}
\usepackage{breqn}
\usepackage{makecell}
\usepackage{multido}
\usepackage{multirow}
\usepackage[separate-uncertainty=true]{siunitx}
\usepackage[capitalize, noabbrev]{cleveref}

\def\eqref#1{equation~\ref{#1}}

\def\floor#1{\lfloor #1 \rfloor}
\def\1{\bm{1}}

\DeclareMathAlphabet{\mathsfit}{\encodingdefault}{\sfdefault}{m}{sl}
\SetMathAlphabet{\mathsfit}{bold}{\encodingdefault}{\sfdefault}{bx}{n}

\def\gG{{\mathcal{G}}}

\def\gL{{\mathcal{L}}}

\def\gN{{\mathcal{N}}}

\def\gX{{\mathcal{X}}}

\def\sN{{\mathbb{N}}}

\newcommand{\E}{\mathbb{E}}

\newcommand{\KL}{D_{\mathrm{KL}}}

\DeclareMathOperator*{\argmax}{arg\,max}

\newcommand{\bs}{\boldsymbol}

\DeclarePairedDelimiter\parentheses{(}{)}

\DeclarePairedDelimiter\brackets{[}{]}
\DeclarePairedDelimiter\lrbrackets{\llbracket}{\rrbracket}
\DeclarePairedDelimiter\rightopeninterval{[}{)}
\DeclarePairedDelimiter\euclideannorm{\|}{\|}
\DeclarePairedDelimiter\leftbracket{[}{.}
\DeclarePairedDelimiter\rightbracket{.}{]}

\DeclarePairedDelimiter\lrfloor{\lfloor}{\rfloor}
\DeclarePairedDelimiterX{\midx}[2]{(}{)}{#1\;\delimsize\vert\;#2}
\DeclarePairedDelimiterX{\parallelx}[2]{(}{)}{#1\;\delimsize\|\;#2}

\def\floor#1{\lrfloor{#1}}

\newcommand{\sMnistTwoImg}[3]{\includegraphics[width=#3\textwidth]{img/samples/mmnist_2d_s_#2/#1.png}}
\newcommand{\bairImg}[3]{\includegraphics[width=#3\textwidth]{img/samples/bair_#2/#1.png}}
\newcommand{\kthImg}[3]{\includegraphics[width=#3\textwidth]{img/samples/kth_#2/#1.png}}

\icmltitlerunning{Stochastic Latent Residual Video Prediction}

\begin{document}

\twocolumn[
    \icmltitle{Stochastic Latent Residual Video Prediction}

    \icmlsetsymbol{equal}{*}

    \begin{icmlauthorlist}
        \icmlauthor{Jean-Yves Franceschi}{equal,lip6}
        \icmlauthor{Edouard Delasalles}{equal,lip6}
        \icmlauthor{Mickaël Chen}{lip6}
        \icmlauthor{Sylvain Lamprier}{lip6}
        \icmlauthor{Patrick Gallinari}{lip6,criteo}
    \end{icmlauthorlist}

    \icmlaffiliation{lip6}{Sorbonne Université, CNRS, LIP6, F-75005 Paris, France}
    \icmlaffiliation{criteo}{Criteo AI Lab, Paris, France}

    \icmlcorrespondingauthor{Jean-Yves Franceschi}{jean-yves.franceschi@lip6.fr}
    \icmlcorrespondingauthor{Edouard Delasalles}{edouard.delasalles@lip6.fr}

    \icmlkeywords{Machine Learning, Deep Learning, Stochastic Video Prediction, Variational Autoencoders, Generative Models, State-Space Models, ICML}

    \vskip 0.3in
]

\renewcommand{\icmlEqualContribution}{\textsuperscript{*}Equal contribution.}
\printAffiliationsAndNotice{\icmlEqualContribution}

\begin{abstract}
    Designing video prediction models that account for the inherent uncertainty of the future is challenging.
    Most works in the literature are based on stochastic image-autoregressive recurrent networks, which raises several performance and applicability issues.
    An alternative is to use fully latent temporal models which untie frame synthesis and temporal dynamics.
    However, no such model for stochastic video prediction has been proposed in the literature yet, due to design and training difficulties.
    In this paper, we overcome these difficulties by introducing a novel stochastic temporal model whose dynamics are governed in a latent space by a residual update rule.
    This first-order scheme is motivated by discretization schemes of differential equations.
    It naturally models video dynamics as it allows our simpler, more interpretable, latent model to outperform prior state-of-the-art methods on challenging datasets.
\end{abstract}

\section{Introduction}

Being able to predict the future of a video from a few conditioning frames in a self-supervised manner has many applications in fields such as reinforcement learning \citep{Gregor2019} or robotics \citep{Babaeizadeh2018}.
More generally, it challenges the ability of a model to capture visual and dynamic representations of the world.
Video prediction has received a lot of attention from the computer vision community. 
However, most proposed methods are deterministic, reducing their ability to capture video dynamics, which are intrinsically stochastic \citep{Denton2018}.

Stochastic video prediction is a challenging task which has been tackled by recent works.
Most state-of-the-art approaches are based on image-autoregressive models \citep{Denton2018, Babaeizadeh2018}, built around Recurrent Neural Networks (RNNs), where each generated frame is fed back to the model to produce the next frame.
However, performances of their temporal models innately depend on the capacity of their encoder and decoder, as each generated frame has to be re-encoded in a latent space.
Such autoregressive processes induce a high computational cost, and strongly tie the frame synthesis and temporal models, which may hurt the performance of the generation process and limit its applicability \citep{Gregor2019, Rubanova2019}.

An alternative approach consists in separating the dynamic of the state representations from the generated frames, which are independently decoded from the latent space.
In addition to removing the aforementioned link between frame synthesis and temporal dynamics, this is computationally appealing when coupled with a low-dimensional latent space.
Moreover, such models can be used to shape a complete representation of the state of a system, e.g. for reinforcement learning applications \citep{Gregor2019}, and are more interpretable than autoregressive models \citep{Rubanova2019}.
Yet, these State-Space Models (SSMs) are more difficult to train as they require non-trivial inference schemes \citep{Krishnan2017} and a careful design of the dynamic model \citep{Karl2017}.
This leads most successful SSMs to only be evaluated on small or artificial toy tasks.

In this work, we introduce a novel stochastic dynamic model for the task of video prediction which successfully leverages structural and computational advantages of SSMs that operate on low-dimensional latent spaces.
Its dynamic component determines the temporal evolution of the system through residual updates of the latent state, conditioned on learned stochastic variables.
This formulation allows us to implement an efficient training strategy and process in an interpretable manner complex high-dimensional data such as videos.
This residual principle can be linked to recent advances relating residual networks and Ordinary Differential Equations (ODEs) \citep{Chen2018}.
This interpretation opens new perspectives such as generating videos at different frame rates, as demonstrated in our experiments.
The proposed approach outperforms current state-of-the-art models on the task of stochastic video prediction, as demonstrated by comparisons with competitive baselines on representative benchmarks.

\section{Related Work}
\label{sec:RelatedWork}

Video synthesis covers a range of different tasks, such as video-to-video translation \citep{Wang2018}, super-resolution \citep{Caballero2017}, interpolation between distant frames \citep{Jiang2018}, generation \citep{Tulyakov2018}, and video prediction, which is the focus of this paper.

\paragraph{Deterministic models.}

Inspired by prior sequence generation models using RNNs \citep{Graves2013}, a number of video prediction methods \citep{Srivastava2015, Villegas2017, Steenkiste2018, Wichers2018, Jin2020} rely on LSTMs \citep[Long Short-Term Memory networks,][]{Hochreiter1997}, or, like \citet{Ranzato2014}, \citet{Jia2016} and \citet{Xu2018a}, on derived networks such as ConvLSTMs \citep{Shi2015}.
Indeed, computer vision approaches are usually tailored to high-dimensional video sequences and propose domain-specific techniques such as pixel-level transformations and optical flow \citep{Shi2015, Walker2015, Finn2016, Jia2016, Walker2016, Vondrick2017, Liang2017, Liu2017, Lotter2017, Lu2017a, Fan2019, Gao2019} that help to produce high-quality predictions.
Such predictions are, however, deterministic, thus hurting their performance as they fail to generate sharp long-term video frames \citep{Babaeizadeh2018, Denton2018}.
Following \citet{Mathieu2016}, some works proposed to use adversarial losses \citep{Goodfellow2014} on the model predictions to sharpen the generated frames \citep{Vondrick2017, Liang2017, Lu2017a, Xu2018b, Wu2020}.
Nonetheless, adversarial losses are notoriously hard to train \citep{Goodfellow2016}, and lead to mode collapse, thereby preventing diversity of generations.

\paragraph{Stochastic and image-autoregressive models.}

Some approaches rely on exact likelihood maximization, using pixel-level autoregressive generation \citep{Oord2016, Kalchbrenner2017, Weissenborn2020} or normalizing flows through invertible transformations between the observation space and a latent space \citep{Kingma2018, Kumar2020}.
However, they require careful design of complex temporal generation schemes manipulating high-dimensional data, thus inducing a prohibitive temporal generation cost.
More efficient continuous models rely on Variational Auto-Encoders \citep[VAEs,][]{Kingma2014, Rezende2014} for the inference of low-dimensional latent state variables.
Except \citet{Xue2016} and \citet{Liu2018} who learn a one-frame-ahead VAE, they model sequence stochasticity by incorporating a random latent variable per frame into a deterministic RNN-based image-autoregressive model.
\citet{Babaeizadeh2018} integrate stochastic variables into the ConvLSTM architecture of \citet{Finn2016}.
Concurrently with \citet{He2018}, \citet{Denton2018} use a prior LSTM conditioned on previously generated frames in order to sample random variables that are fed to a predictor LSTM; performance of such methods were improved in follow-up works by increasing networks capacities \citep{Castrejon2019, Villegas2019}.
Finally, \citet{Lee2018} combine the ConvLSTM architecture and this learned prior, adding an adversarial loss on the predicted videos to sharpen them at the cost of a diversity drop.
Yet, all these methods are image-autoregressive, as they feed their predictions back into the latent space, thereby tying the frame synthesis and temporal models and increasing their computational cost.
Concurrently to our work, \citet{Minderer2019} propose to use the autoregressive VRNN model \citep{Chung2015} on learned image key-points instead of raw frames.
It remains unclear to which extent this change could mitigate the aforementioned problems.
We instead tackle these issues by focusing on video dynamics, and propose a model that is state-space and acts on a small latent space.
This approach yields better experimental results despite weaker video-specific priors.

\paragraph{State-space models.}

Many latent state-space models have been proposed for sequence modelization \citep{Bayer2014, Fraccaro2016, Fraccaro2017, Krishnan2017, Karl2017, Hafner2019}, usually trained by deep variational inference.
These methods, which use locally linear or RNN-based dynamics, are designed for low-dimensional data, as learning such models on complex data is challenging, or focus on control or planning tasks.
In contrast, our fully latent method is the first one to be successfully applied to complex high-dimensional data such as videos, thanks to a temporal model based on residual updates of its latent state.
It falls within the scope of a recent trend linking differential equations with neural networks \citep{Lu2017b, Long2018}, leading to the integration of ODEs, that are seen as continuous residual networks \citep{He2016}, in neural network architectures \citep{Chen2018}.
However, the latter work as well as follow-ups and related works \citep{Rubanova2019, Yildiz2019, LeGuen2020} are either limited to low-dimensional data, prone to overfitting or unable to handle stochasticity within a sequence.
Another line of works considers stochastic differential equations with neural networks \citep{Ryder2018, DeBrouwer2019}, but are limited to continuous Brownian noise, whereas video prediction additionally requires to model punctual stochastic events.

\section{Model}
\label{sec:Model}

\begin{figure*}
    \begin{minipage}[c]{0.206\textwidth}
        \centering
        \subfigure[Generative model $p$.]{\label{fig:Model}\large\resizebox{\columnwidth}{!}{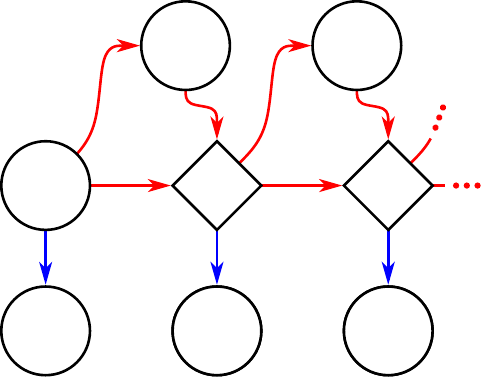}}
        \par
        \subfigure[Inference model $q$.]{\label{fig:Inference}\large\resizebox{\columnwidth}{!}{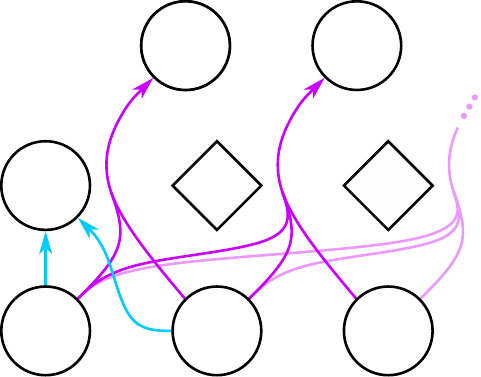}}
    \end{minipage}%
    \hfill
    \begin{minipage}[c]{.77\textwidth}
        \centering
        \subfigure[
            Model and inference architecture on a test sequence.
            The transparent block on the left depicts the prior, and those on the right correspond to the full inference performed at training time.
        ]{\label{fig:Architecture}\large\resizebox{\columnwidth}{!}{\scriptsize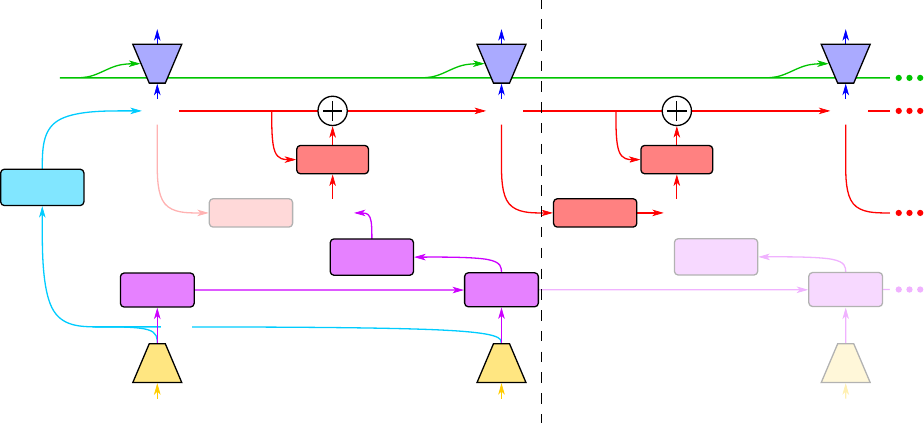}}
    \end{minipage}%
    \vspace{-0.03in}
    \caption{
        (a), (b) Proposed generative and inference models.
        Diamonds and circles represent, respectively, deterministic and stochastic states.
        (c) Corresponding architecture with two parts: inference on conditioning frames on the left, generation for extrapolation on the right.
        $h_{\phi}$ and $g_{\theta}$ are deep Convolutional Neural Networks (CNNs), and other named networks are Multilayer Perceptrons (MLPs).
    }
    \vspace{-0.13in}
\end{figure*}

We consider the task of stochastic video prediction, consisting in approaching, given a number of conditioning video frames, the distribution of possible future frames.

\subsection{Latent Residual Dynamic Model}

Let $\bs{x}_{1:T}$ be a sequence of $T$ video frames.
We model their evolution by introducing latent variables $\bs{y}$ that are driven by a dynamic temporal model.
Each frame $\bs{x}_{t}$ is then generated from the corresponding latent state $\bs{y}_{t}$ only, making the dynamics independent from the previously generated frames.

We propose to model the transition function of the latent dynamic of $\bs{y}$ with a stochastic residual network.
State $\bs{y}_{t + 1}$ is chosen to deterministically depend on the previous state $\bs{y}_{t}$, conditionally to an auxiliary random variable $\bs{z}_{t + 1}$.
These auxiliary variables encapsulate the randomness of the video dynamics.
They have a learned factorized Gaussian prior that depends on the previous state only.
The model is depicted in \cref{fig:Model}, and defined as follows:
\begin{equation}
    \label{eq:DynamicModel}
    \begin{cases}
        \bs{y}_{1} \sim \gN\parentheses{\bs{0}, I}, \\
        \bs{z}_{t + 1} \sim \gN\parentheses*{\mu_{\theta} \parentheses*{\bs{y}_{t}}, \sigma_{\theta}\parentheses*{\bs{y}_{t}} I}, \\
        \bs{y}_{t + 1} = \bs{y}_{t} + f_{\theta}\parentheses*{\bs{y}_{t}, \bs{z}_{t + 1}}, \\
        \bs{x}_{t} \sim \gG\parentheses*{g_{\theta} \parentheses*{\bs{y}_{t}}},
    \end{cases}
\end{equation}
where $\mu_\theta$, $\sigma_\theta$, $f_\theta$ and $g_{\theta}$ are neural networks, and $\gG\parentheses*{g_{\theta} \parentheses*{\bs{y}_{t}}}$ is a probability distribution parameterized by $g_{\theta}\parentheses*{\bs{y}_{t}}$.
In our experiments, $\gG$ is a normal distribution with mean $g_{\theta}\parentheses*{\bs{y}_{t}}$ and constant diagonal variance.
Note that $\bs{y}_{1}$ is assumed to have a standard Gaussian prior, and, in our VAE setting, will be inferred from conditioning frames for the prediction task, as shown in \cref{sec:Inference}.

The residual update rule takes inspiration in the Euler discretization scheme of differential equations.
The state of the system $\bs{y}_{t}$ is updated by its first-order movement, i.e., the residual $f_{\theta} \parentheses*{\bs{y}_{t}, \bs{z}_{t + 1}}$.
Compared to a regular RNN, this simple principle makes our temporal model lighter and more interpretable.
\cref{eq:DynamicModel}, however, differs from a discretized ODE because of the introduction of the stochastic discrete-time variables $\bs{z}$.
Nonetheless, we propose to allow the Euler step size $\Delta t$ to be smaller than $1$, as a way to make the temporal model closer to a continuous dynamics.
The updated dynamics becomes, with $\frac{1}{\Delta t} \in \sN$ to synchronize the step size with the video frame rate:
\begin{equation}
    \label{eq:IncreasedFramerate}
    \bs{y}_{t + \Delta t} = \bs{y}_{t} + \Delta t \cdot f_{\theta} \parentheses*{\bs{y}_{t}, \bs{z}_{\floor{t} + 1}}.
\end{equation}
For this formulation, the auxiliary variable $\bs{z}_t$ is kept constant between two integer time steps.
Note that a different $\Delta t$ can be used during training or testing.
This allows our model to generate videos at an arbitrary frame rate since each intermediate latent state can be decoded in the observation space.
This ability enables us to observe the quality of the learned dynamic as well as challenge its ODE inspiration by testing its generalization to the continuous limit in \cref{sec:Experiments}.
In the following, we consider $\Delta t$ as a hyperparameter.
For the sake of clarity, we consider that $\Delta t = 1$ in the remaining of this section; generalizing to a smaller $\Delta t$ is straightforward as \cref{fig:Model} remains unchanged.

\subsection{Content Variable}
\label{sec:ContentVariable}

Some components of video sequences can be static, such as the background or shapes of moving objects.
They may not impact the dynamics; we therefore model them separately, in the same spirit as \citet{Denton2017} and \citet{Yingzhen2018}. We compute a content variable $\bs{w}$ that remains constant throughout the whole generation process and is fed together with $\bs{y}_{t}$ into the frame generator.
It enables the dynamical part of the model to focus only on movement, hence being lighter and more stable.
Moreover, it allows us to leverage architectural advances in neural networks, such as skip connections \citep{Ronneberger2015}, to produce more realistic frames.

This content variable is a deterministic function $c_{\psi}$ of a fixed number $k < T$ of frames $\bs{x}_{\mathrm{c}}^{\parentheses*{k}} = \parentheses*{\bs{x}_{i_{1}}, \ldots, \bs{x}_{i_{k}}}$:
\begin{equation}
    \label{eq:ContentVariable}
    \begin{cases}
        \bs{w} = c_{\psi}\parentheses*{\bs{x}_{\mathrm{c}}^{\parentheses*{k}}} = c_{\psi}\parentheses*{\bs{x}_{i_{1}}, \ldots, \bs{x}_{i_{k}}} \\
        \bs{x}_{t} \sim \gG\parentheses*{g_{\theta}\parentheses*{\bs{y}_{t}, \bs{w}}}.
    \end{cases}
\end{equation}
During testing, $\bs{x}_{\mathrm{c}}^{\parentheses*{k}}$ are the last $k$ conditioning frames (usually between $2$ and $5$).

This content variable is not endowed with any probabilistic prior, contrary to the dynamic variables $\bs{y}$ and $\bs{z}$.
Thus, the information it contains is not constrained in the loss function (see \cref{sec:Inference}), but only architecturally.
To prevent temporal information from leaking in $\bs{w}$, we propose to uniformly sample these $k$ frames within $\bs{x}_{1:T}$ during training.
We also design $c_{\psi}$ as a permutation-invariant function \citep{Zaheer2017}, consisting in an MLP fed with the sum of individual frame representations, following \citet{Santoro2017}.

This absence of prior and its architectural constraint allows $\bs{w}$ to contain as much non-temporal information as possible, while preventing it from containing dynamic information.
On the other hand, due to their strong standard Gaussian priors, $\bs{y}$ and $\bs{z}$ are encouraged to discard unnecessary information.
Therefore, $\bs{y}$ and $\bs{z}$ should only contain temporal information that could not be captured by $\bs{w}$.

Note that this content variable can be removed from our model, yielding a more classical deep state-space model. An experiment in this setting is presented in \cref{app:Pendulum}.

\subsection{Variational Inference and Architecture}
\label{sec:Inference}

Following the generative process depicted in \cref{fig:Model}, the conditional joint probability of the full model, given a content variable $\bs{w}$, can be written as:
\begin{equation}
    \label{eq:Joint}
    \begin{multlined}
        p\midx*{\bs{x}_{1:T}, \bs{z}_{2:T}, \bs{y}_{1:T}}{\bs{w}} \\
        = p\parentheses*{\bs{y}_{1}} \prod_{t = 2}^{T} p\midx*{\bs{z}_{t}, \bs{y}_{t}}{\bs{y}_{t - 1}} \prod_{t = 1}^T p\midx*{\bs{x}_{t}}{\bs{y}_t, \bs{w}},
    \end{multlined}
\end{equation}
with
\begin{equation}
    \label{eq:JointTemporal}
    p\midx*{\bs{z}_{t}, \bs{y}_{t}}{\bs{y}_{t - 1}} = p\midx*{\bs{z}_{t}}{\bs{y}_{t - 1}} p\midx*{\bs{y}_{t}}{\bs{y}_{t - 1}, \bs{z}_{t}}.
\end{equation}
According to the expression of $\bs{y}_{t + 1}$ in \cref{eq:DynamicModel}, $p\midx*{\bs{y}_{t}}{\bs{y}_{t - 1}, \bs{z}_{t}} = \delta\parentheses*{\bs{y}_{t} - \bs{y}_{t - 1} - f_{\theta} \parentheses*{\bs{y}_{t - 1}, \bs{z}_{t}}}$, where $\delta$ is the Dirac delta function centered on $\bs{0}$.
Hence, in order to optimize the likelihood of the observed videos $p\midx*{\bs{x}_{1:T}}{\bs{w}}$, we need to infer latent variables $\bs{y}_{1}$ and $\bs{z}_{2:T}$.
This is done by deep variational inference using the inference model parameterized by $\phi$ and shown in \cref{fig:Inference}, which comes down to considering a variational distribution $q_{Z,Y}$ defined and factorized as follows:
\begin{equation}
    \label{eq:JointQ}
    \begin{multlined}
        q_{Z,Y} \triangleq q\midx*{\bs{z}_{2:T},\bs{y}_{1:T}}{\bs{x}_{1:T}, \bs{w}} \\
        = q\midx*{\bs{y}_{1}}{\bs{x}_{1:k}} \prod_{t = 2}^{T} q\midx*{\bs{z}_{t}}{\bs{x}_{1:t}} \underbrace{q\midx*{\bs{y}_{t}}{\bs{y}_{t - 1}, \bs{z}_{t}}}_{= p\midx*{\bs{y}_{t}}{\bs{y}_{t - 1}, \bs{z}_{t}}},
    \end{multlined}
\end{equation}
with $q\midx*{\bs{y}_{t}}{\bs{y}_{t - 1}, \bs{z}_{t}} = p\midx*{\bs{y}_{t}}{\bs{y}_{t - 1}, \bs{z}_{t}}$ being the aforementioned Dirac delta function.
This yields the following evidence lower bound (ELBO), whose full derivation is given in \cref{app:ELBO}:
\begin{equation}
\label{eq:ELBO}
    \begin{aligned}
    \MoveEqLeft[2] \log p\midx*{\bs{x}_{1:T}}{\bs{w}} \geq \gL\parentheses*{\bs{x}_{1:T}; \bs{w}, \theta, \phi} & \\
    \triangleq & - \KL\parallelx*{q\midx*{\bs{y}_{1}}{\bs{x}_{1:k}}}{p\parentheses*{\bs{y}_1}} \\
    & + \E_{\parentheses*{\widetilde{\bs{z}}_{2:T}, \widetilde{\bs{y}}_{1:T}} \sim q_{Z,Y}} \leftbracket*{\sum_{t=1}^{T} \log p\midx*{\bs{x}_{t}}{\widetilde{\bs{y}}_{t}, \bs{w}}} \\
    & \rightbracket*{\mathop{\hphantom{+}} - \sum_{t=2}^{T} \KL\parallelx*{q\midx*{\bs{z}_{t}}{\bs{x}_{1: t}}}{p\midx*{\bs{z}_{t}}{\widetilde{\bs{y}}_{t - 1}}}},
    \end{aligned}
\end{equation}
where $\KL$ denotes the Kullback–Leibler (KL) divergence \citep{Kullback1951}.

The sum of KL divergence expectations implies to consider the full past sequence of inferred states for each time step, due to the dependence on conditionally deterministic variables $\bs{y}_{2:T}$.
However, optimizing $\gL\parentheses*{\bs{x}_{1:T}; \bs{w}, \theta, \phi}$ with respect to model parameters $\theta$ and variational parameters $\phi$ can be done efficiently by sampling a single full sequence of states from $q_{Z,Y}$ per example, and computing gradients by backpropagation \citep{Rumelhart1988} trough all inferred variables, using the reparameterization trick \citep{Kingma2014, Rezende2014}.
We classically choose $q\midx*{\bs{y}_{1}}{\bs{x}_{1:k}}$ and $q\midx*{\bs{z}_{t}}{\bs{x}_{1:t}}$ to be factorized Gaussian so that all KL divergences can be computed analytically.

We include an $\ell_2$ regularization term on residuals $f_{\theta}$ applied to $\bs{y}$ which stabilizes the temporal dynamics of the residual network, as noted by \citet{Behrmann2019}, \citet{deBezenac2019} and \citet{Rousseau2019}.
Given a set of videos $\gX$, the full optimization problem, where $\gL$ is defined as in \cref{eq:ELBO}, is then given as:
\begin{equation}
    \label{eq:Loss}
    \begin{multlined}
        \argmax_{\theta,\phi,\psi} \sum_{\bs{x} \in \gX} \leftbracket*{\mathop{\vphantom{\sum_{t=2}^{T}}} \E_{\bs{x}_{\mathrm{c}}^{\parentheses*{k}}} \gL\parentheses*{\bs{x}_{1:T}; c_{\psi}\parentheses*{\bs{x}_{\mathrm{c}}^{\parentheses*{k}}}, \theta, \phi}} \\
        \rightbracket*{- \lambda \cdot \E_{(\bs{z}_{2:T}, \bs{y}_{1:T}) \sim q_{Z,Y}} \sum_{t=2}^{T} \euclideannorm*{f_{\theta} \parentheses*{\bs{y}_{t-1}, \bs{z}_{t}}}_2}.
    \end{multlined}
\end{equation}

\begin{figure*}
    \centering
    \includegraphics[width=\textwidth]{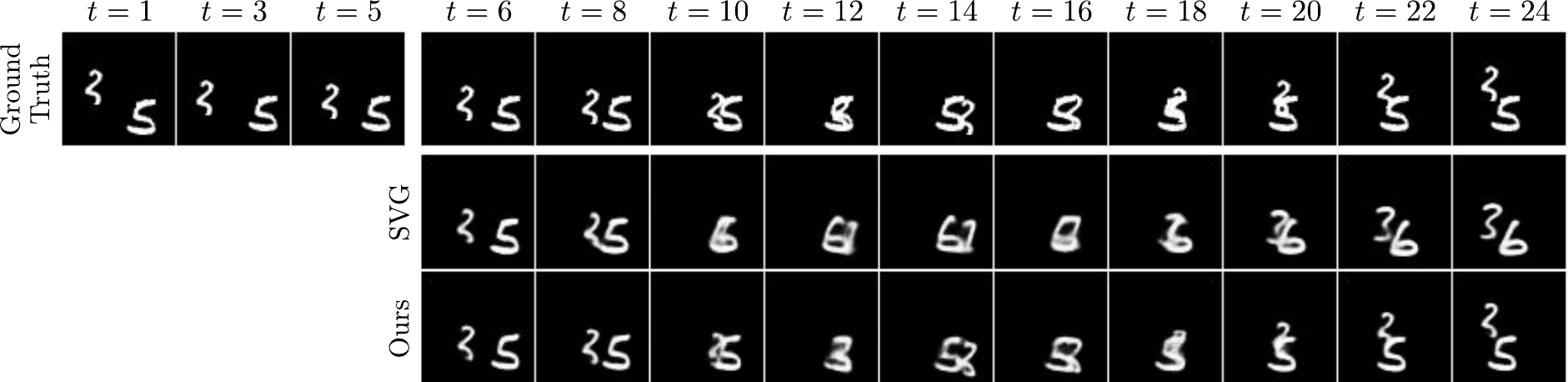}
    \vspace{-0.2in}
    \caption{
        \label{fig:mmnist-2d-s-sample-1}
        Conditioning frames and corresponding ground truth and best samples with respect to PSNR from SVG and our method for an example of the Stochastic Moving MNIST dataset.
    }
    \vspace{-0.13in}
\end{figure*}

\cref{fig:Architecture} depicts the full architecture of our temporal model, showing how the model is applied during testing.
The first latent variables are inferred with the conditioning framed and are then predicted with the dynamic model.
In contrast, during training, each frame of the input sequence is considered for inference, which is done as follows.
Firstly, each frame $\bs{x}_{t}$ is independently encoded into a vector-valued representation $\widetilde{\bs{x}}_{t}$, with $\widetilde{\bs{x}}_{t} = h_{\phi}\parentheses*{\bs{x}_{t}}$.
$\bs{y}_{1}$ is then inferred using an MLP on the first $k$ encoded frames $\widetilde{\bs{x}}_{1: k}$.
Each $\bs{z}_{t}$ is inferred in a feed-forward fashion with an LSTM on the encoded frames.
Inferring $\bs{z}$ this way experimentally performs better than, e.g., inferring them from the whole sequence $\bs{x}_{1: T}$; we hypothesize that this follows from the fact that this filtering scheme is closer to the prediction setting, where the future is not available.

\section{Experiments}
\label{sec:Experiments}

This section exposes the experimental results of our method on four standard stochastic video prediction datasets.\footnote{Code and datasets are available at \url{https://github.com/edouardelasalles/srvp}. Pretrained models are downloadable at \url{https://data.lip6.fr/srvp/}.}
We compare our method with state-of-the-art baselines on stochastic video prediction.
Furthermore, we qualitatively study the dynamics and latent space learned by our model.\footnote{Animated video samples are available at \url{https://sites.google.com/view/srvp/}.}
Training details are described in \cref{app:Training}.

\begin{figure*}
    \centering
    \includegraphics[width=\textwidth]{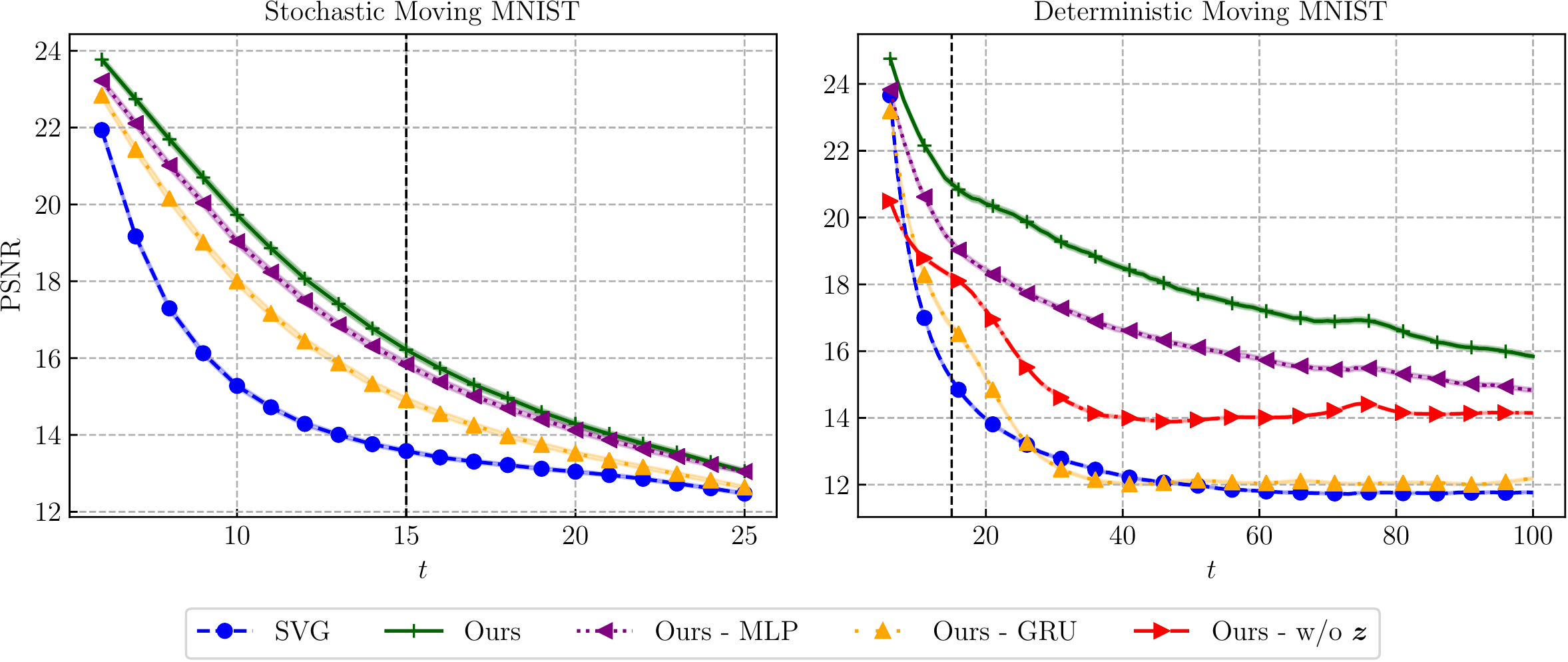}
    \vspace{-0.2in}
    \caption{
        \label{fig:res-mmnist-2d}
        Mean PSNR scores with respect to $t$ for all tested models on the Moving MNIST dataset, with their $95\%$-confidence intervals.
        Vertical bars mark the length of training sequences.
    }
\end{figure*}

\begin{figure*}
    \centering
    \includegraphics[width=\textwidth]{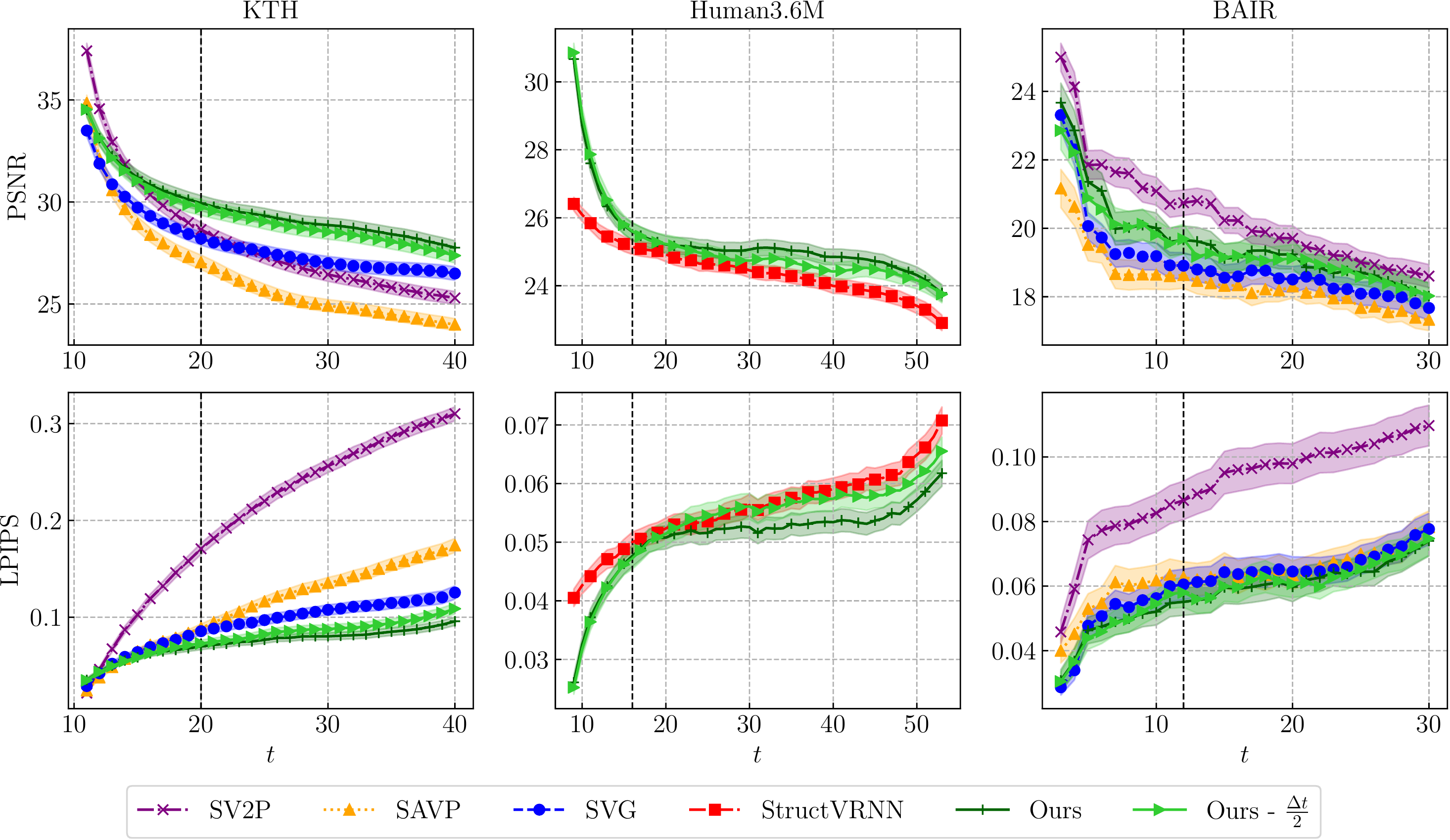}
    \vspace{-0.2in}
    \caption{
        \label{fig:res-kth-human-bair}
        PSNR and LPIPS scores with respect to $t$ for all tested models on the KTH (left column), Human3.6M (center) and BAIR (right) datasets, with their $95\%$-confidence intervals.
        Vertical bars mark the length of training sequences.
    }
\end{figure*}

\begin{table*}
    \caption{
        \label{tab:fvd}
        FVD scores for all tested methods on the KTH, Human3.6M and BAIR datasets with their $95\%$-confidence intervals over five different samples from the models.
        Bold scores indicate the best performing method for each dataset.
    }
    \sisetup{detect-weight, table-align-uncertainty=true, mode=text}
    \renewrobustcmd{\bfseries}{\fontseries{b}\selectfont}
    \renewrobustcmd{\boldmath}{}
    \centering
    \vspace{0.1in}
    \small
    \begin{tabular}{lS[table-format=3(2)]S[table-format=3(2)]S[table-format=3(2)]S[table-format=3(2)]S[table-format=3(2)]S[table-format=3(2)]S[table-format=3(2)]S[table-format=3(2)]S[table-format=3(2)]S[table-format=3(2)]}
        \toprule
        Dataset & {SV2P} & {SAVP} & {SVG} & {StructVRNN} & {Ours} & {Ours - $\frac{\Delta t}{2}$} & {Ours - MLP} & {Ours - GRU}\tabularnewline
        \midrule
        KTH & 636 \pm 1 & 374 \pm 3 & 377 \pm 6 & {\textemdash} & \bfseries 222 \pm 3 & 244 \pm 3 & 255 \pm 4 & 240 \pm 5 \tabularnewline
        Human3.6M & {\textemdash} & {\textemdash} & {\textemdash} & 556 \pm 9 & \bfseries 416 \pm 5 & \bfseries 415 \pm 3 & 582 \pm 4 & 1050 \pm 20 \tabularnewline
        BAIR & 965 \pm 17 & \bfseries 152 \pm 9 & 255 \pm 4 & {\textemdash} & 163 \pm 4 & 222 \pm 42 & 162 \pm 4 & 178 \pm 10 \tabularnewline
        \bottomrule
    \end{tabular}
    \vspace{-0.1in}
\end{table*}

The stochastic nature and novelty of the task of stochastic video prediction make it challenging to evaluate \citep{Lee2018}: since videos and models are stochastic, comparing the ground truth and a predicted video is not adequate.
We thus adopt the common approach \citep{Denton2018, Lee2018} consisting in, for each test sequence, sampling from the tested model a given number (here, $100$) of possible futures and reporting the best performing sample against the true video.
We report this discrepancy for three commonly used metrics that are computed frame-wise and averaged over time: Peak Signal-to-Noise Ratio (PSNR, \emph{higher is better}), Structured Similarity (SSIM, \emph{higher is better}), and Learned Perceptual Image Patch Similarity \citep[LPIPS, \emph{lower is better},][]{Zhang2018}.
PSNR greatly penalizes errors in predicted dynamics, as it is a pixel-level measure derived from the $\ell_{2}$ distance, but might also favor blurry predictions.
SSIM (only reported in \cref{app:Results} for the sake of concision) rather compares local frame patches to circumvent this issue, but loses some dynamics information. 
LPIPS compares images through a learned distance between activations of deep CNNs trained on image classification tasks, and has been shown to better correlate with human judgment on real images.
Finally, the recently proposed Fréchet Video Distance \citep[FVD, \emph{lower is better},][]{Unterthiner2018} aims at directly comparing the distribution of predicted videos with the ground truth distribution through the representations computed by a deep CNN trained on action recognition tasks.
It has been shown, independently from LPIPS, to better capture the realism of predicted videos than PSNR and SSIM.
We treat all four metrics as complementary, as they capture different scales and modalities.

\begin{figure*}
    \centering
    \includegraphics[width=\textwidth]{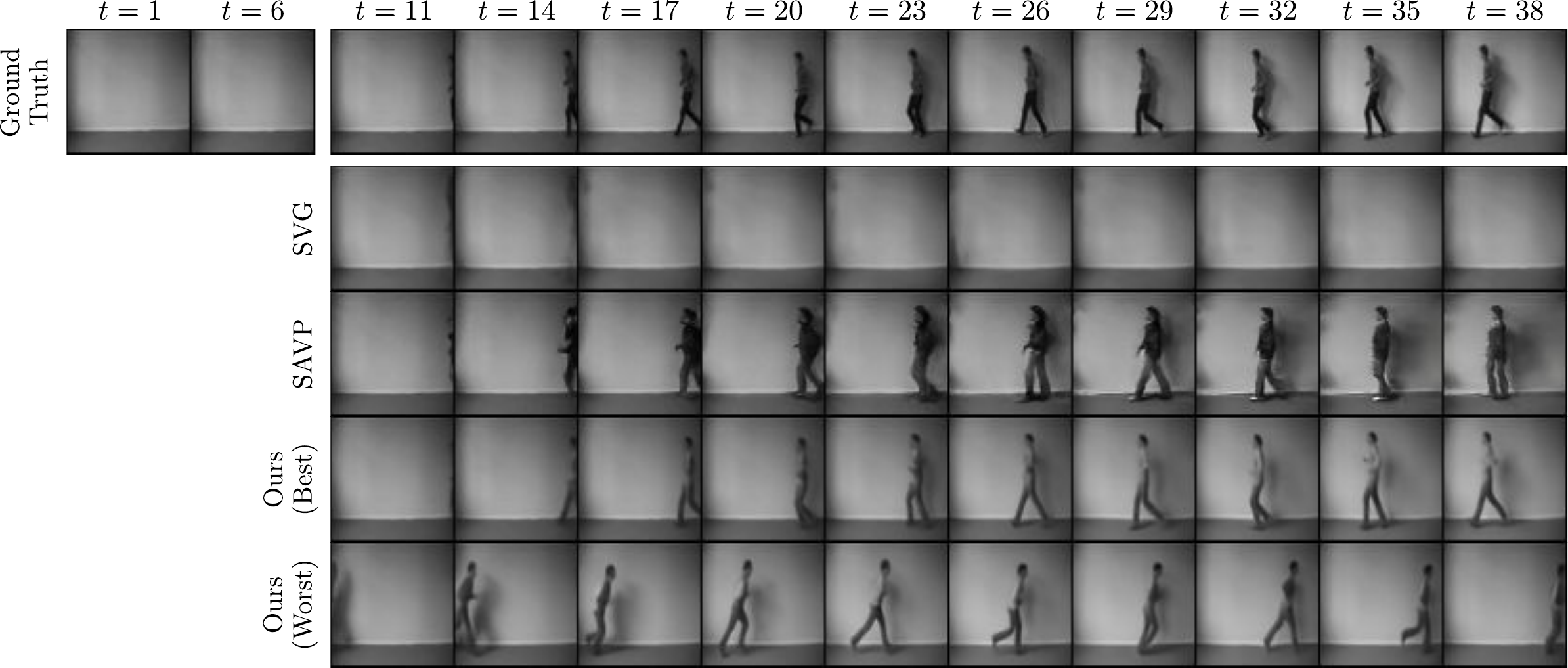}
    \vspace{-0.2in}
    \caption{
        \label{fig:kth-sample-1}
        Conditioning frames and corresponding ground truth, best samples from SVG, SAVP and our method, and worst sample from our method, for a video of the KTH dataset.
        Samples are chosen according to their LPIPS with respect to the ground truth.
        SVG fails to make a person appear, unlike SAVP and our model.
        The latter better predicts the subject pose and produces more realistic predictions.
    }
    \vspace{-0.13in}
\end{figure*}

We present experimental results on a simulated dataset and three real-world datasets, that we briefly present in the following and detail in \cref{app:Dataset}.
The corresponding numerical results can be found in \cref{app:Results}.
For the sake of concision, we only display a handful of qualitative samples in this section, and refer to \cref{app:Samples} and our website for additional samples.
We compare our model against several variational state-of-the-art models: SV2P \citep{Babaeizadeh2018}, SVG \citep{Denton2018}, SAVP \citep{Lee2018}, and StructVRNN \citep{Minderer2019}.
Note that SVG has the closest training and architecture to ours among the state of the art.
Therefore, we use the same neural architecture as SVG for our encoders and decoders in order to perform fair comparisons with this method.

All baseline results are presented only on the datasets on which they were tested in the original articles.
They were obtained with pretrained models released by the authors, except those of SVG on the Moving MNIST dataset and StructVRNN on the Human3.6M dataset, for which we trained models using the code and hyperparameters provided by the authors (see \cref{app:Dataset}).
Unless specified otherwise, our model is tested with the same $\Delta t$ as in training (see \cref{eq:IncreasedFramerate}).

\paragraph{Stochastic Moving MNIST.}

This dataset consists of one or two MNIST digits \citep{LeCun1998} moving linearly and randomly bouncing on walls with new direction and velocity sampled randomly at each bounce \citep{Denton2018}.

\cref{fig:res-mmnist-2d} (left) shows quantitative results with two digits.
Our model outperforms SVG on both PSNR and SSIM; LPIPS and FVD are not reported as they are not relevant for this synthetic task.
Decoupling dynamics from image synthesis allows our method to maintain temporal consistency despite high-uncertainty frames where crossing digits become indistinguishable.
For instance in \cref{fig:mmnist-2d-s-sample-1}, the digits shapes change after they cross in the SVG prediction, while our model predicts the correct digits.
To evaluate the predictive ability on a longer horizon, we perform experiments on the deterministic version of the dataset \citep{Srivastava2015} with only one prediction per model to compute PSNR and SSIM.
We show the results up to $t + 95$ in \cref{fig:res-mmnist-2d} (right).
We can see that our model better captures the dynamics of the problem compared to SVG as its performance decreases significantly less over time, especially at a long-term horizon.

We also compare to two alternative versions of our model in \cref{fig:res-mmnist-2d}, where the residual dynamic function is replaced by an MLP or a GRU \citep[Gated Recurrent Unit,][]{Cho2014}.
Our residual model outperforms both versions on the stochastic, and especially on the deterministic version of the dataset, showing its intrinsic advantage at modeling long-term dynamics.
Finally, on the deterministic version of Moving MNIST, we compare to an alternative where $\bs{z}$ is entirely removed, resulting in a temporal model close to the one presented by \citet{Chen2018}.
The loss of performance of this alternative model is significant, showing that our stochastic residual model offers a substantial advantage even when used in a deterministic environment.

\begin{figure*}
    \centering
    \includegraphics[width=0.87\textwidth]{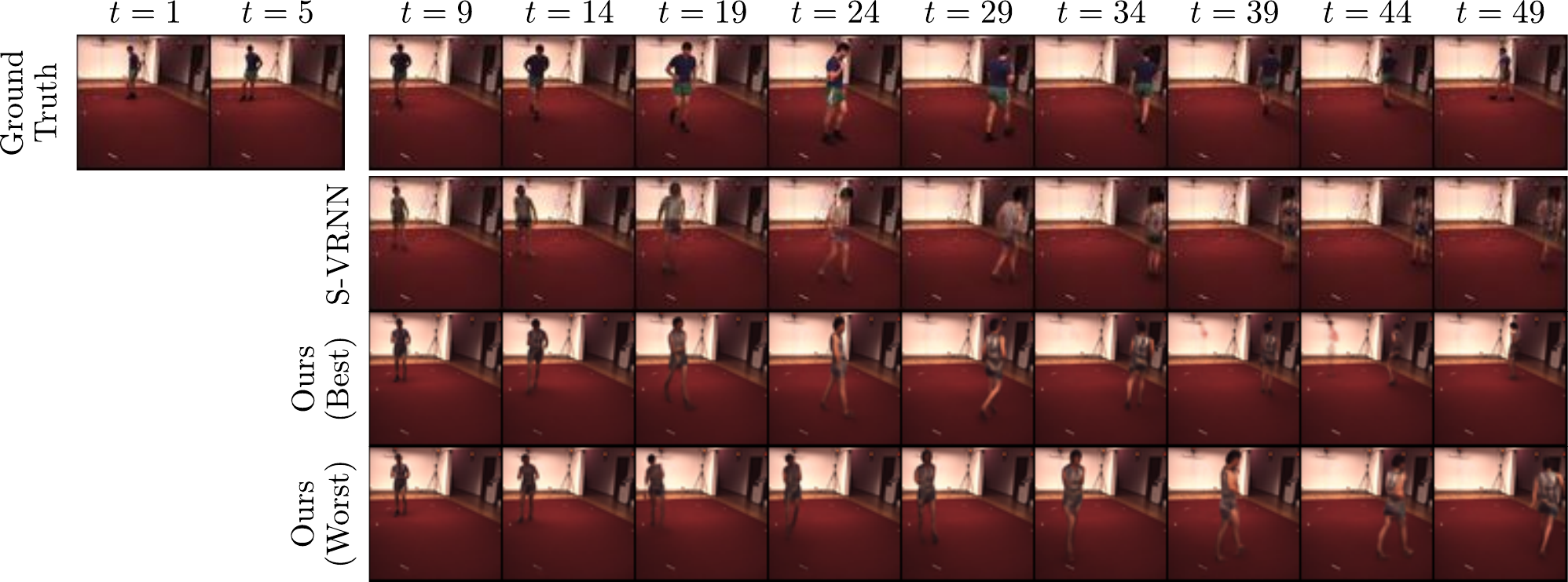}
    \vspace{-0.03in}
    \caption{
        \label{fig:human-sample-1}
        Conditioning frames and corresponding ground truth, best samples from StructVRNN and our method, and worst sample from our method, with respect to LPIPS, for a video of the Human3.6M dataset.
        Our method better captures the dynamic of the subject and produces less artefacts than StructVRNN.
    }
    \vspace{-0.13in}
\end{figure*}

\paragraph{KTH Action dataset (KTH).}

This dataset is composed of real-world videos of people performing a single action per video in front of different backgrounds \citep{Schuldt2004}.
Uncertainty lies in the appearance of subjects, the actions they perform, and how they are performed.

We substantially outperform on this dataset every considered baseline for each metric, as shown in \cref{fig:res-kth-human-bair,tab:fvd}.
In some videos, the subject only appears after the conditioning frames, requiring the model to sample the moment and location of the subject appearance, as well as its action.
This critical case is illustrated in \cref{fig:kth-sample-1}.
There, SVG fails to even generate a moving person; only SAVP and our model manage to do so, and our best sample is closer to the subject's poses compared to SAVP.
Moreover, the worst sample of our model demonstrates that it captures the diversity of the dataset by making a person appear at different time steps and with different speeds.
An additional experiment on this dataset in \cref{app:Autoregressivity} studies the influence of the encoder and decoder architecture on SVG and our model.

Finally, \cref{tab:fvd} and appendix \cref{tab:res-kth} compare our method to its MLP and GRU alternative versions, leading to two conclusions.
Firstly, it confirms the structural advantage of residual dynamics observed on Moving MNIST.
Indeed, both MLP and GRU lose on all metrics, and especially in terms of realism according to LPIPS and FVD.
Secondly, all three versions of our model (residual, MLP, GRU) outperform prior methods. 
Therefore, this improvement is due to their common inference method, latent nature and content variable, strengthening our motivation to propose a non-autoregressive model.

\paragraph{Human3.6M.}

This dataset is also made of videos of subjects performing various actions \citep{Ionescu2011, Ionescu2014}.
While there are more actions and details to capture with less training subjects than in KTH, the video backgrounds are less varied, and subjects always remain within the frames.

As reported in \cref{fig:res-kth-human-bair,tab:fvd}, we significantly outperform, with respect to all considered metrics, StructVRNN, which is the state of the art on this dataset and has been shown to surpass both SAVP and SVG by \citet{Minderer2019}.
\cref{fig:human-sample-1} shows the dataset challenges; in particular, both methods do not capture well the subject appearance.
Nonetheless, our model better captures its movements, and produces more realistic frames.

Comparisons to the MLP and GRU versions demonstrate once again the advantage of using residual dynamics.
GRU obtains low scores on all metrics, which is coherent with similar results for SVG reported by \citet{Minderer2019}.
While the MLP version remains close to the residual model on PSNR, SSIM and LPIPS, it is largely beaten by the latter in terms of FVD.

\begin{figure}
    \centering
    \subfigure[Cropped KTH sample.]{\kthImg{hyp_2_dt}{702}{0.43}}
    \subfigure[Cropped Human3.6M sample.]{\includegraphics[width=0.43\textwidth]{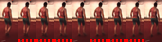}}
    \subfigure[Cropped BAIR sample.]{\bairImg{hyp_2_dt}{155}{0.43}}
    
    \vspace{-0.03in}
    \caption{
        \label{fig:oversampling-1}
        Generation examples at doubled or quadrupled frame rate, using a halved $\Delta t$ compared to training.
        Frames including a bottom red dashed bar are intermediate frames.
    }
    \vspace{-0.13in}
\end{figure}

\begin{figure}
    \centering
    \includegraphics[width=\columnwidth]{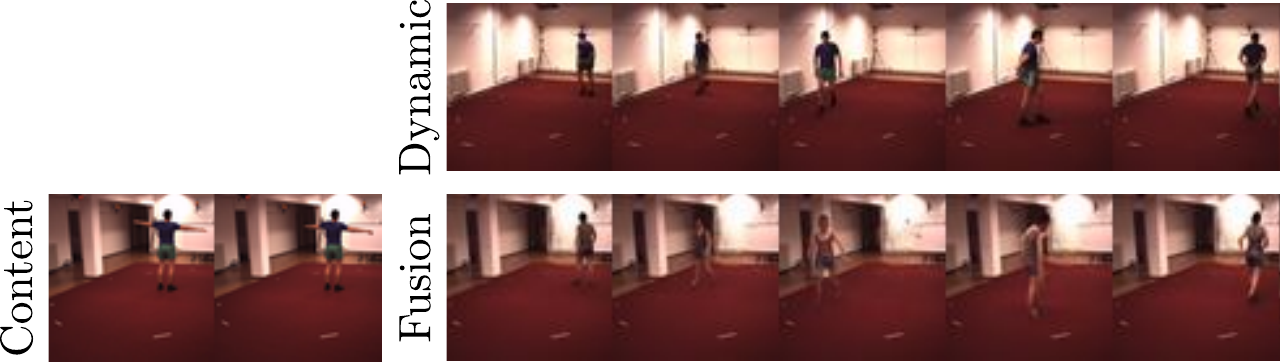}
    \vspace{-0.2in}
    \caption{
        \label{fig:human-content-swap-1}
        Video (bottom right) generated from the dynamic latent state $\bs{y}$ inferred with a video (top) and the content variable $\bs{w}$ computed with the conditioning frames of another video (left).
        The generated video keeps the same background as the bottom left frames, while the subject moves accordingly to the top frames.
    }
    \vspace{-0.13in}
\end{figure}

\begin{figure*}
    \centering
    \includegraphics[width=\textwidth]{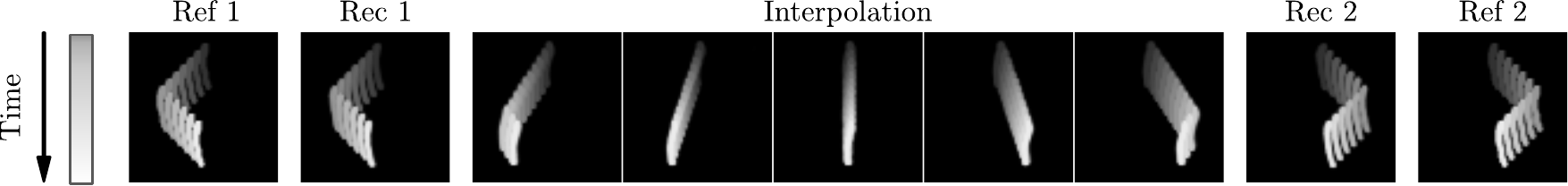}
    \vspace{-0.2in}
    \caption{
        \label{fig:mnist-interpolation-1}
        From left to right, $\bs{x}^{\mathrm{s}}$, $\widehat{\bs{x}}^{\mathrm{s}}$ (reconstruction of $\bs{x}^{\mathrm{s}}$ by the VAE of our model), results of the interpolation in the latent space between $\bs{x}^{\mathrm{s}}$ and $\bs{x}^{\mathrm{t}}$, $\widehat{\bs{x}}^{\mathrm{t}}$ and $\bs{x}^{\mathrm{t}}$.
        Each trajectory is materialized in shades of grey in the frames.
    }
    \vspace{-0.03in}
\end{figure*}

\paragraph{BAIR robot pushing dataset (BAIR).}

This dataset contains videos of a Sawyer robotic arm pushing objects on a tabletop \citep{Ebert2017}.
It is highly stochastic as the arm can change its direction at any moment.
Our method achieves similar or better results compared to state-of-the-art models in terms of PSNR, SSIM and LPIPS, as shown in \cref{fig:res-kth-human-bair}, except for SV2P that produces very blurry samples, as seen in \cref{app:Samples}, yielding good PSNR but prohibitive LPIPS scores.
Our method obtains second-best FVD score, close to SAVP whose adversarial loss enables it to better model small objects, and outperforms SVG, whose variational architecture is closest to ours, demonstrating the advantage of non-autoregressive methods.
Recent advances \citep{Villegas2019} indicate that performance of such variational models can be improved by increasing networks capacities, but this is out of the scope of this paper.

\paragraph{Varying frame rate in testing.}

We challenge here the ODE inspiration of our model.
\cref{eq:IncreasedFramerate} amounts to learning a residual function $f_{\bs{z}_{\floor{t} + 1}}$ over $t \in \rightopeninterval*{\floor{t}, \floor{t} + 1}$.
We aim at testing whether this dynamics is close to its continuous generalization:
\begin{equation}
    \od{\bs{y}}{t} = f_{\bs{z}_{\floor{t} + 1}}\parentheses*{\bs{y}},
\end{equation}
which is a piecewise ODE.
To this end, we refine this Euler approximation during testing by halving $\Delta t$; if this maintains the performance of our model, then the dynamic rule of the latter is close to the piecewise ODE.
As shown in \cref{fig:res-kth-human-bair,tab:fvd}, prediction performances overall remain stable while generating twice as many frames (cf. \cref{app:Oversampling} for further discussion).
Therefore, the justification of the proposed update rule is supported by empirical evidence.
This property can be used to generate videos at a higher frame rate, with the same model, and without supervision.
We show in \cref{fig:oversampling-1} and \cref{app:Oversampling} frames generated at a double and quadruple frame rate on KTH, Human3.6M and BAIR.

\paragraph{Disentangling dynamics and content.}

Let us show that the proposed model actually separates content from dynamics as discussed in \cref{sec:ContentVariable}.
To this end, two sequences $\bs{x}^{\mathrm{s}}$ and $\bs{x}^{\mathrm{t}}$ are drawn from the Human3.6M testing set.
While $\bs{x}^{\mathrm{s}}$ is used for extracting our content variable $\bs{w}^{\mathrm{s}}$, dynamic states $\bs{y}^{\mathrm{t}}$ are inferred with our model from  $\bs{x}^{\mathrm{t}}$.
New frame sequences $\widehat{\bs{x}}$ are finally generated from the fusion of the content vector and the dynamics.
This results in a content corresponding to the first sequence $\bs{x}^{\mathrm{s}}$ and a movement following the dynamics of the second sequence $\bs{x}^{\mathrm{t}}$, as observed in \cref{fig:human-content-swap-1}.
More samples for KTH, Human3.6M, and BAIR can be seen in \cref{app:Samples}.

\paragraph{Interpolation of dynamics.}

Our state-space structure allows us to learn semantic representations in $\bs{y}_{t}$.
To highlight this feature, we test whether two deterministic Moving MNIST trajectories can be interpolated by linearly interpolating their inferred latent initial conditions.
We begin by generating two trajectories $\bs{x}^{\mathrm{s}}$ and $\bs{x}^{\mathrm{t}}$ of a single moving digit.
We infer their respective latent initial conditions $\bs{y}_{1}^{\mathrm{s}}$ and $\bs{y}_{1}^{\mathrm{t}}$.
We then use our model to generate frame sequences from latent initial conditions linearly interpolated between $\bs{y}_{1}^{\mathrm{s}}$ and $\bs{y}_{1}^{\mathrm{t}}$.
If it learned a meaningful latent space, the resulting trajectory should also be a smooth interpolation between the directions of reference trajectories $\bs{x}^{\mathrm{s}}$ and $\bs{x}^{\mathrm{t}}$, and this is what we observe in \cref{fig:mnist-interpolation-1}. 
Additional examples can be found in \cref{app:Samples}.

\section{Conclusion}

We introduce a novel dynamic latent model for stochastic video prediction which, unlike prior image-autoregressive models, decouples frame synthesis and dynamics.
This temporal model is based on residual updates of a small latent state that is showed to perform better than RNN-based models.
This endows our method with several desirable properties, such as temporal efficiency and latent space interpretability.
We experimentally demonstrate the performance and advantages of the proposed model, which outperforms prior state-of-the-art methods for stochastic video prediction.
This work is, to the best of our knowledge, the first to propose a latent dynamic model that scales for video prediction.
The proposed model is also novel with respect to the recent line of work dealing with neural networks and ODEs for temporal modeling; it is the first such residual model to scale to complex stochastic data such as videos.

We believe that the general principles of our model (state-space, residual dynamic, static content variable) can be generally applied to other models as well.
Interesting future works include replacing the VRNN of \citet{Minderer2019} with our residual dynamics in order to model the evolution of key-points, supplementing our model with more video-specific priors, or leveraging its state-space nature in model-based reinforcement learning.

\section*{Acknowledgements}

We would like to thank all members of the MLIA team from the LIP6 laboratory of Sorbonne Université for helpful discussions and comments, as well as Matthias Minderer for his help to process the Human3.6M dataset and reproduce StructVRNN results.

We acknowledge financial support from the LOCUST ANR project (ANR-15-CE23-0027).
This work was granted access to the HPC resources of IDRIS under the allocation 2020-AD011011360 made by GENCI (Grand Equipement National de Calcul Intensif).

\bibliography{refs}
\bibliographystyle{template/icml2020}

\onecolumn

\appendix

\section{Evidence Lower Bound}
\label{app:ELBO}

We develop in this section the computations of the variational lower bound for the proposed model.

Using the original variational lower bound of \citet{Kingma2014} in \cref{eq:OriginalELBO}:
\begingroup
\allowdisplaybreaks
\begin{align}
    \nonumber \MoveEqLeft[3] \log p\midx*{\bs{x}_{1: T}}{\bs{w}} & \\
    \label{eq:OriginalELBO}
    \geq {} & \E_{\parentheses*{\widetilde{\bs{z}}_{2:T}, \widetilde{\bs{y}}_{1:T}} \sim q_{Z,Y}} \log p\midx*{\bs{x}_{1:T}}{\widetilde{\bs{z}}_{2:T}, \widetilde{\bs{y}}_{1:T}, \bs{w}} - \KL\parallelx*{q_{Z,Y}}{p\midx*{\bs{y}_{1:T}, \bs{z}_{2:T}}{\bs{w}}} \\
    \begin{split}
        \label{eq:ModelDependencies}
        = {} & \E_{\parentheses*{\widetilde{\bs{z}}_{2:T}, \widetilde{\bs{y}}_{1:T}} \sim q_{Z,Y}} \log p\midx*{\bs{x}_{1:T}}{\widetilde{\bs{z}}_{2:T}, \widetilde{\bs{y}}_{1:T}, \bs{w}} - \KL\parallelx*{q\midx*{\bs{y}_{1}, \bs{z}_{2:T}}{\bs{x}_{1:T}}}{p\parentheses*{\bs{y}_{1}, \bs{z}_{2:T}}}
    \end{split} \\
    \label{eq:MutualIndependenceX}
    = {} & \E_{\parentheses*{\widetilde{\bs{z}}_{2:T}, \widetilde{\bs{y}}_{1:T}} \sim q_{Z,Y}} \sum_{t = 1}^{T} \log p\midx*{\bs{x}_{t}}{\widetilde{\bs{y}}_{t}, \bs{w}} - \KL\parallelx*{q\midx*{\bs{y}_{1}, \bs{z}_{2:T}}{\bs{x}_{1:T}}}{p\parentheses*{\bs{y}_{1}, \bs{z}_{2:T}}}, \\
    \intertext{
        where:
        \begin{itemize}
            \item \cref{eq:ModelDependencies} is given by the forward and inference models factorizing $p$ and $q$ in \cref{eq:Joint,eq:JointTemporal,eq:JointQ} and illustrated by, respectively, \cref{fig:Model,fig:Inference}:
            \begin{itemize}
                \item the $\bs{z}$ variables and $\bs{y}_{1}$ are independent from $\bs{w}$ with respect to $p$ and $q$;
                \item the $\bs{y}_{2:T}$ variables are deterministic functions of $\bs{y}_{1}$ and $\bs{z}_{2:T}$ with respect to $p$ and $q$;
            \end{itemize}
            \item \cref{eq:MutualIndependenceX} results from the factorization of $p\midx*{\bs{x}_{1:T}}{\bs{y}_{1:T}, \bs{z}_{1:T}, \bs{w}}$ in \cref{eq:Joint}.
        \end{itemize}
        From there, by using the integral formulation of $\KL$:
    }
    \nonumber \MoveEqLeft[3] \log p\midx*{\bs{x}_{1: T}}{\bs{w}} & \\
    \begin{split}
        \geq {} & \E_{\parentheses*{\widetilde{\bs{z}}_{2:T}, \widetilde{\bs{y}}_{1:T}} \sim q_{Z,Y}} \sum_{t = 1}^{T} \log p\midx*{\bs{x}_{t}}{\widetilde{\bs{y}}_{t}, \bs{w}} \\
        & + \idotsint_{\bs{y}_{1}, \bs{z}_{2:T}} q\midx*{\bs{y}_{1}, \bs{z}_{2:T}}{\bs{x}_{1:T}} \log \frac{p\parentheses*{\bs{y}_{1}, \bs{z}_{2:T}}}{q\midx*{\bs{y}_{1}, \bs{z}_{2:T}}{\bs{x}_{1:T}}} \dif \bs{z}_{2:T} \dif \bs{y}_{1}
    \end{split} \\
    \begin{split}
        = {} & \E_{\parentheses*{\widetilde{\bs{z}}_{2:T}, \widetilde{\bs{y}}_{1:T}} \sim q_{Z,Y}} \sum_{t = 1}^{T} \log p\midx*{\bs{x}_{t}}{\widetilde{\bs{y}}_{t}, \bs{w}} - \KL\parallelx*{q\midx*{\bs{y}_{1}}{\bs{x}_{1:T}}}{p\parentheses*{\bs{y}_{1}}} \\
        & + \E_{\widetilde{\bs{y}}_{1} \sim q\midx*{\bs{y}_{1}}{\bs{x}_{1:T}}} \brackets*{\idotsint_{\bs{z}_{2:T}} q\midx*{\bs{z}_{2:T}}{\bs{x}_{1:T}, \widetilde{\bs{y}}_{1}} \log \frac{p\midx*{\bs{z}_{2:T}}{\widetilde{\bs{y}}_{1}}}{q\midx*{\bs{z}_{2:T}}{\bs{x}_{1:T}, \widetilde{\bs{y}}_{1}}} \dif \bs{z}_{2:T}}
    \end{split} \\
    \begin{split}
        \label{eq:ModelDependencies2}
        = {} & \E_{\parentheses*{\widetilde{\bs{z}}_{2:T}, \widetilde{\bs{y}}_{1:T}} \sim q_{Z,Y}} \sum_{t = 1}^{T} \log p\midx*{\bs{x}_{t}}{\widetilde{\bs{y}}_{t}, \bs{w}} - \KL\parallelx*{q\midx*{\bs{y}_{1}}{\bs{x}_{1:k}}}{p\parentheses*{\bs{y}_{1}}} \\
        & + \E_{\widetilde{\bs{y}}_{1} \sim q\midx*{\bs{y}_{1}}{\bs{x}_{1:k}}} \brackets*{\idotsint_{\bs{z}_{2:T}} q\midx*{\bs{z}_{2:T}}{\bs{x}_{1:T}, \widetilde{\bs{y}}_{1}} \log \frac{p\midx*{\bs{z}_{2:T}}{\widetilde{\bs{y}}_{1}}}{q\midx*{\bs{z}_{2:T}}{\bs{x}_{1:T}, \widetilde{\bs{y}}_{1}}} \dif \bs{z}_{2:T}}
    \end{split} \\
    \begin{split}
        \label{eq:TemporalFactorization}
        = {} & \E_{\parentheses*{\widetilde{\bs{z}}_{2:T}, \widetilde{\bs{y}}_{1:T}} \sim q_{Z,Y}} \sum_{t = 1}^{T} \log p\midx*{\bs{x}_{t}}{\widetilde{\bs{y}}_{t}, \bs{w}} - \KL\parallelx*{q\midx*{\bs{y}_{1}}{\bs{x}_{1:k}}}{p\parentheses*{\bs{y}_{1}}} \\
        & + \E_{\widetilde{\bs{y}}_{1} \sim q\midx*{\bs{y}_{1}}{\bs{x}_{1:k}}} \brackets*{\idotsint_{\bs{z}_{2:T}} \prod_{t=2}^{T} q\midx*{\bs{z}_{t}}{\bs{x}_{1:t}} \sum_{t=2}^{T} \log \frac{p\midx*{\bs{z}_{t}}{\widetilde{\bs{y}}_{1}, \bs{z}_{2:t - 1}}}{q\midx*{\bs{z}_{t}}{\bs{x}_{1:t}}} \dif \bs{z}_{2:T}}
    \end{split} \\
    \begin{split}
        \label{eq:FirstIterationELBO}
        = {} & \E_{\parentheses*{\widetilde{\bs{z}}_{2:T}, \widetilde{\bs{y}}_{1:T}} \sim q_{Z,Y}} \sum_{t = 1}^{T} \log p\midx*{\bs{x}_{t}}{\widetilde{\bs{y}}_{t}, \bs{w}} - \KL\parallelx*{q\midx*{\bs{y}_{1}}{\bs{x}_{1:k}}}{p\parentheses*{\bs{y}_{1}}} \\
        & - \E_{\widetilde{\bs{y}}_{1} \sim q\midx*{\bs{y}_{1}}{\bs{x}_{1:k}}} \KL\parallelx*{q\midx*{\bs{z}_{2}}{\bs{x}_{1:t}}}{p\midx*{\bs{z}_{2}}{\widetilde{\bs{y}}_{1}}} \\
        & + \E_{\widetilde{\bs{y}}_{1} \sim q\midx*{\bs{y}_{1}}{\bs{x}_{1:k}}} \E_{\widetilde{\bs{z}}_{2} \sim q\midx*{\bs{z}_{2}}{\bs{x}_{1:2}}} \brackets*{\idotsint_{\bs{z}_{3:T}} \prod_{t=3}^{T} q\midx*{\bs{z}_{t}}{\bs{x}_{1:t}} \sum_{t=3}^{T} \log \frac{p\midx*{\bs{z}_{t}}{\bs{y}_{1}, \widetilde{\bs{z}}_{2:t - 1}}}{q\midx*{\bs{z}_{t}}{\bs{x}_{1:t}}} \dif \bs{z}_{3:T}},
    \end{split} \\
    \intertext{
        where:
        \begin{itemize}
            \item \cref{eq:ModelDependencies2} follows from the inference model of \cref{eq:JointQ}, where $\bs{y}_{1}$ only depends on $\bs{x}_{1:k}$;
            \item \cref{eq:TemporalFactorization} is obtained from the factorizations of \cref{eq:Joint,eq:JointTemporal,eq:JointQ}.
        \end{itemize}
        By iterating \cref{eq:FirstIterationELBO}'s step on $\bs{z}_{3}, \ldots, \bs{z}_{T}$ and factorizing all expectations, we obtain:
    }
    \nonumber \MoveEqLeft[3] \log p\midx*{\bs{x}_{1: T}}{\bs{w}} & \\
    \begin{split}
        \geq {} & \E_{\parentheses*{\widetilde{\bs{z}}_{2:T}, \widetilde{\bs{y}}_{1:T}} \sim q_{Z,Y}} \sum_{t = 1}^{T} \log p\midx*{\bs{x}_{t}}{\widetilde{\bs{y}}_{t}, \bs{w}} - \KL\parallelx*{q\midx*{\bs{y}_{1}}{\bs{x}_{1:k}}}{p\parentheses*{\bs{y}_{1}}} \\
        & - \E_{\widetilde{\bs{y}}_{1} \sim q\midx*{\bs{y}_{1}}{\bs{x}_{\mathrm{c}}}} \parentheses*{\E_{\widetilde{\bs{z}}_{t} \sim q\midx*{\bs{z}_{t}}{\bs{x}_{1:t}}}}_{t = 2}^{T} \sum_{t=2}^{T} \KL\parallelx*{q\midx*{\bs{z}_{t}}{\bs{x}_{1: t}}}{p\midx*{\bs{z}_{t}}{\widetilde{\bs{y}}_{1}, \widetilde{\bs{z}}_{1:t-1}}},
    \end{split} \\
    \intertext{and we finally retrieve \cref{eq:ELBO} by using the factorization of \cref{eq:JointQ}:}
    \nonumber \MoveEqLeft[3] \log p\midx*{\bs{x}_{1: T}}{\bs{w}} & \\
    \begin{split}
        \geq {} & \E_{\parentheses*{\widetilde{\bs{z}}_{2:T}, \widetilde{\bs{y}}_{1:T}} \sim q_{Z,Y}} \sum_{t = 1}^{T} \log p\midx*{\bs{x}_{t}}{\widetilde{\bs{y}}_{t}, \bs{w}} - \KL\parallelx*{q\midx*{\bs{y}_{1}}{\bs{x}_{1:k}}}{p\parentheses*{\bs{y}_{1}}} \\
        & - \E_{\parentheses*{\widetilde{\bs{z}}_{2:T}, \widetilde{\bs{y}}_{1:T}} \sim q_{Z,Y}} \sum_{t=2}^{T} \KL\parallelx*{q\midx*{\bs{z}_{t}}{\bs{x}_{1: t}}}{p\midx*{\bs{z}_{t}}{\widetilde{\bs{y}}_{t - 1}}}.
    \end{split}
\end{align}
\endgroup

\section{Datasets Details}
\label{app:Dataset}

We detail in this section the datasets used in our experimental study.

\subsection{Data Representation}

For all datasets, video frames are represented by greyscale or RGB pixels with values within $\brackets*{0, 1}$ obtained by dividing by $255$ their original values lying in $\lrbrackets*{0, 255}$.

\subsection{Stochastic Moving MNIST}
\label{app:Dataset-SMMNIST}

This monochrome dataset consists in one or two training MNIST digits \citep{LeCun1998} of size $28 \times 28$ moving linearly within a $64 \times 64$ frame and randomly bouncing against its border, sampling a new direction and velocity at each bounce \citep{Denton2018}.
We use the same settings as \citet{Denton2018}, train all models on $15$ timesteps, condition them at testing time on $5$ frames, and predict either $20$ (for the stochastic version) or $95$ (for the deterministic version) frames.
Note that we adapted the dataset to sample more coherent bounces: the original dataset computes digit trajectories that are dependent on the chosen framerate, unlike our corrected version of the dataset.
We consequently retrained SVG on this dataset, obtaining comparable results as those originally presented by \citet{Denton2018}.
Test data were produced by generating a trajectory for each testing digit, and randomly pairwise combining these trajectories to produce \num{5000} testing sequences containing each two digits.

\subsection{KTH Action Dataset (KTH)}
\label{app:Dataset-KTH}

This dataset is composed of real-world $64 \times 64$ monochrome videos of $25$ people performing one of six actions (walking, jogging, running, boxing, handwaving and handclapping) in front of different backgrounds \citep{Schuldt2004}.
Uncertainty lies in the appearance of subjects, the action they perform and how it is performed.
We use the same settings as \citet{Denton2018}, train all models on $20$ timesteps, condition them at testing time on $10$ frames, and predict $30$ frames.
The training set is formed with actions from the first $20$ subjects, the remaining five being used for testing.
Training is performed by sampling sub-sequences of size $20$ in the training set.
The test set is composed of \num{1000} randomly sampled sub-sequences of size $40$.

\subsection{Human3.6M}
\label{app:Dataset-Human}

This dataset is also made of videos of subjects performing various actions \citep{Ionescu2011, Ionescu2014}.
While there are more actions and details to capture with less training subjects than in KTH, the video backgrounds are less varied, and subjects always remain within the frames.
We use the same settings as \citet{Minderer2019} to train both our model and StructVRNN, for which there is no available pretrained model.
We train all models on $16$ timesteps, condition them at testing time on $8$ frames, and predict $45$ frames.
Videos used in our experiment are subsampled from the original videos at $6.25$Hz, center-cropped from $1000 \times 1000$ to $800 \times 800$ and resized to $64 \times 64$ using the Lanczos filter of the Pillow library\footnote{\url{https://pillow.readthedocs.io/}}.
The training set is composed of videos of subjects $1$, $5$, $6$, $7$, and $8$, and the testing set is made from subjects $9$ and $11$; videos showing more than one action, marked by ``ALL'' in the dataset, are excluded.
Training is performed by sampling sub-sequences of size $16$ in the training set.
The test set is composed of \num{1000} randomly sampled sub-sequences of size $53$ from the testing videos.

\subsection{BAIR Robot Pushing Dataset (BAIR)}
\label{app:Dataset-BAIR}

This dataset contains $64 \times 64$ videos of a Sawyer robotic arm pushing objects on a tabletop \citep{Ebert2017}.
It is highly stochastic as the arm can change its direction at any moment.
We use the same settings as \citet{Denton2018}, train all models on $12$ timesteps, condition them at testing time on $2$ frames, and predict $28$ frames.
Training and testing sets are the same as those used by \citet{Denton2018}.

\section{Training Details}
\label{app:Training}

We expose in this section further information needed for the reproduction of our results.

\subsection{Specifications}

We used Python $3.7.6$ and PyTorch $1.4.0$ \citep{Paszke2019} to implement our model.
Each model was trained on Nvidia GPUs with CUDA $10.1$ using mixed-precision training \citep{Micikevicius2018} with Apex.\footnote{\url{https://github.com/nvidia/apex}.}

\subsection{Architecture}

\paragraph{Encoder and decoder architecture.}

Both $g_{\theta}$ and $h_{\phi}$ are chosen to have the same mirrored architecture that depends on the dataset.
We used the same architectures as \citet{Denton2018}: a DCGAN discriminator and generator architecture \citep{Radford2016} for Moving MNIST, and a VGG16 \citep{Simonyan2015} architecture (mirrored for $h_{\phi}$) for the other datasets.
In both cases, the output of $h_{\phi}$ (i.e., $\widetilde{\bs{x}}$) is a vector of size $128$, and $g_{\theta}$ and $h_{\phi}$ weights are initialized using a centered Gaussian distribution with a standard deviation of $0.02$ (except for biases initialized to $0$, and batch normalization layers weights drawn from a Gaussian distribution with unit mean and a standard deviation of $0.02$).
Additionally, we supplement $g_{\theta}$ with a last sigmoid activation in order to ensure its outputs lie within $\brackets*{0, 1}$ like the ground truth data.

Note that, during testing, predicted frames are directly generated by $g_{\theta}\parentheses*{\bs{y}_{t}, \bs{w}}$ without sampling from the observation probability distribution $\gG\parentheses*{g_{\theta}\parentheses*{\bs{y}_{t}, \bs{w}}} = \gN\parentheses*{g_{\theta}\parentheses*{\bs{y}_{t}, \bs{w}}, \nu I}$.
This is a common practice for Gaussian decoders in VAEs that is adopted by our competitors \citep{Lee2018, Denton2018, Minderer2019}.

\paragraph{Content variable.}
For the Moving MNIST dataset, the content variable $\bs{w}$ is obtained directly from $\widetilde{\bs{x}}$ and is a vector of size $128$.
For KTH, Human3.6M, and BAIR, we supplement this vectorial variable with skip connections from all layers of the encoder $g_{\theta}$ that are then fed to the decoder $h_{\phi}$ to handle complex backgrounds.
For Moving MNIST, the number of frames $k$ used to compute the content variable is $5$; for KTH and Human3.6M, it is $3$; for BAIR, it is $2$.

The vectorial content variable $\bs{w}$ is computed from $k$ input frames $\bs{x}_{\mathrm{c}}^{\parentheses*{k}} = \parentheses*{\bs{x}_{i_{1}}, \ldots, \bs{x}_{i_{k}}}$ with $c_{\psi}$ defined as follows:
\begin{equation}
    \bs{w} = c_{\psi}\parentheses*{\bs{x}_{\mathrm{c}}^{\parentheses*{k}}} = c_{\psi}^{2}\parentheses*{\sum_{j = 1}^{k} c_{\psi}^{1}\parentheses*{\widetilde{\bs{x}}_{i_{j}}}}.
\end{equation}
In other words, $c_{\psi}$ transforms each frame representation using $c_{\psi}^{1}$, sums these transformations and outputs the application of $c_{\psi}^{2}$ to this sum.
Since frame representations $\widetilde{\bs{x}}_{i_{j}} = h_{\phi}\parentheses*{\bs{x}_{i_{j}}}$ are computed independently from each other, $c_{\psi}$ is indeed permutation-invariant.
In practice, $c_{\psi}^{1}$ consists in a linear layer of output size $256$ followed by a rectified linear unit (ReLU) activation, while $c_{\psi}^{2}$ is a linear layer of output size $256$ (making $\bs{w}$ of size $256$) followed by a hyperbolic tangent activation.

\paragraph{LSTM architecture.}

The LSTM used for all datasets has a single layer of LSTM cells with a hidden state size of $256$.

\paragraph{MLP architecture.}

All MLPs used in inference (with parameters $\phi$) have three linear layers with hidden size $256$ and ReLU activations.
All MLPs used in the forward model (with parameters $\theta$) have four linear layers with hidden size 512 and ReLU activations.
Any MLP outputting Gaussian distribution parameters $\parentheses*{\mu, \sigma}$ additionally includes a softplus \citep{Dugas2001} applied to its output dimensions that are used to obtain $\sigma$.
Weights of $f_{\theta}$ are orthogonally initialized with a gain of $1.2$ for KTH and Human3.6M, and $1.41$ for the other datasets (except for biases which are initialized to $0$), while the other MLPs are initialized with the default weight initialization of PyTorch.

\paragraph{Sizes of latent variables.}

The sizes of the latent variables in our model are the following: for Moving MNIST, $\bs{y}$ and $\bs{z}$ have size $20$;  for KTH, Human3.6M, and BAIR, $\bs{y}$ and $\bs{z}$ have size $50$.

\paragraph{Euler step size}

Models are trained with $\Delta t = 1$ on Moving MNIST, and with $\Delta t = \frac{1}{2}$ on the others datasets.

\subsection{Optimization}

Models are trained using the Adam optimizer \citep{Kingma2015} with learning rate \num{3e-4}, and decay rates $\beta_1 = 0.9$ and $\beta_2 = 0.999$.

\paragraph{Loss function.}

The batch size is chosen to be $128$ for Moving MNIST, $100$ for KTH and Human3.6M, and $192$ for BAIR.
The regularization coefficient $\lambda$ is always set to $1$.
Logarithms used in the loss are natural logarithms.

For the Moving MNIST dataset, we follow \citet{Higgins2017} by weighting the KL divergence terms on $\bs{z}$ (i.e., the sum of KL divergences in \cref{eq:ELBO}) with a multiplication factor $\beta = 2$.

\paragraph{Variance of the observation.}

The variance $\nu$ considered in the observation probability distribution $\gG\parentheses*{g_{\theta}\parentheses*{\bs{y}, \bs{w}}} = \gN\parentheses*{g_{\theta}\parentheses*{\bs{y}_{t}, \bs{w}}, \nu I}$ is chosen as follows:
\begin{itemize}
    \item for Moving MNIST, $\nu = 1$;
    \item for KTH and Human3.6M, $\nu = $ \num{4e-2};
    \item for BAIR, $\nu = \frac{1}{2}$.
\end{itemize}

\paragraph{Number of optimization steps.}

The number of optimization steps for each dataset is the following:
\begin{itemize}
    \item Stochastic Moving MNIST: \num{1000000} steps, with additional \num{100000} steps where the learning rate is linearly decreased to $0$;
    \item Deterministic Moving MNIST: \num{800000} steps, with additional \num{100000} steps where the learning rate is linearly decreased to $0$;
    \item KTH: \num{150000} steps, with additional \num{50000} steps where the learning rate is linearly decreased to $0$;
    \item Human3.6M: \num{325000} steps, with additional \num{25000} steps where the learning rate is linearly decreased to $0$;
    \item BAIR: \num{1000000} steps, with additional \num{500000} steps where the learning rate is linearly decreased to $0$.
\end{itemize}

Furthermore, the final models for KTH and Human3.6M are chosen among several checkpoints, computed every \num{5000} iterations for KTH and \num{20000} iterations for Human3.6M, as the ones obtaining the best evaluation PSNR.
This evaluation score differs from the test score as we extract from the training set an evaluation set by randomly selecting $5\%$ of the training videos from the training set of each dataset.
More precisely, the evaluation PSNR for a checkpoint is computed as the mean best prediction PSNR for $400$ (for KTH) or $200$ (for Human3.6M) randomly extracted sequences of length $30$ (for KTH) or $53$ (for Human3.6M) from the videos of the evaluation set.

\section{Additional Numerical Results}
\label{app:Results}

\begin{table}
    \caption{
        \label{tab:res-smmnist}
        Numerical results (mean and $95\%$-confidence interval) for PSNR and SSIM of tested methods on the two-digits Moving MNIST dataset.
        Bold scores indicate the best performing method for each metric and, where appropriate, scores whose means lie in the confidence interval of the best performing method.
    }
    \sisetup{detect-weight, table-align-uncertainty=true, mode=text}
    \renewrobustcmd{\bfseries}{\fontseries{b}\selectfont}
    \renewrobustcmd{\boldmath}{}
    \centering
    \vspace{0.1in}
    \begin{tabular}{lS[table-format=2.2(2)]S[table-format=1.4(2)]S[table-format=2.2(2)]S[table-format=1.4(2)]}
        \toprule
        \multicolumn{1}{l}{\multirow{2}{*}[-0.5\dimexpr \aboverulesep + \belowrulesep + \cmidrulewidth]{Models}} & \multicolumn{2}{c}{Stochastic} & \multicolumn{2}{c}{Deterministic} \tabularnewline
        \cmidrule(lr){2-3}\cmidrule(lr){4-5}
        {} & {PSNR} & {SSIM} & {PSNR} & {SSIM} \tabularnewline
        \midrule
        SVG & 14.50 \pm 0.04 & 0.7090 \pm 0.0015 & 12.85 \pm 0.03 & 0.6185 \pm 0.0011 \tabularnewline
        Ours & \bfseries 16.93 \pm 0.07 & \bfseries 0.7799 \pm 0.0020 & \bfseries 18.25 \pm 0.06 & \bfseries 0.8300 \pm 0.0017 \tabularnewline
        Ours - MLP & 16.55 \pm 0.06 & 0.7694 \pm 0.0019 & 16.70 \pm 0.05 & 0.7876 \pm 0.0015 \tabularnewline
        Ours - GRU & 15.80 \pm 0.05 & 0.7464 \pm 0.0016 & 13.17 \pm 0.03 & 0.6237 \pm 0.0011 \tabularnewline
        Ours - w/o $\bs{z}$ & {\textemdash} & {\textemdash} & 14.99 \pm 0.03 & 0.4757 \pm 0.0019 \tabularnewline
        \bottomrule
    \end{tabular}
\end{table}

\begin{table}
    \caption{
        \label{tab:res-kth}
        Numerical results (mean and $95\%$-confidence interval, when relevant) for PSNR, SSIM, and LPIPS of tested methods on the KTH dataset.
        Bold scores indicate the best performing method for each metric and, where appropriate, scores whose means lie in the confidence interval of the best performing method.
    }
    \sisetup{detect-weight, table-align-uncertainty=true, mode=text}
    \renewrobustcmd{\bfseries}{\fontseries{b}\selectfont}
    \renewrobustcmd{\boldmath}{}
    \centering
    \vspace{0.1in}
    \begin{tabular}{lS[table-format=2.2(2)]S[table-format=1.4(2)]S[table-format=1.4(2)]}
        \toprule
        Models & {PSNR} & {SSIM} & {LPIPS} \tabularnewline
        \midrule
        SV2P & 28.19 \pm 0.31 & 0.8141 \pm 0.0050 & 0.2049 \pm 0.0053 \tabularnewline
        SAVP & 26.51 \pm 0.29 & 0.7564 \pm 0.0062 & 0.1120 \pm 0.0039 \tabularnewline
        SVG & 28.06 \pm 0.29 & 0.8438 \pm 0.0054 & 0.0923 \pm 0.0038 \tabularnewline
        Ours & \bfseries 29.69 \pm 0.32 & \bfseries 0.8697 \pm 0.0046 & \bfseries 0.0736 \pm 0.0029 \tabularnewline
        Ours - $\frac{\Delta t}{2}$ & \bfseries 29.43 \pm 0.33 & 0.8633 \pm 0.0049 & 0.0790 \pm 0.0034 \tabularnewline
        Ours - MLP & 28.91 \pm 0.34 & 0.8527 \pm 0.0051 & 0.0799 \pm 0.0032 \tabularnewline
        Ours - GRU & 29.14 \pm 0.33 & 0.8590 \pm 0.0050 & 0.0790 \pm 0.0032 \tabularnewline
        \bottomrule
    \end{tabular}
\end{table}

\begin{table}
    \caption{
        \label{tab:res-human}
        Numerical results (mean and $95\%$-confidence interval, when relevant) for PSNR, SSIM, and LPIPS of tested methods on the Human3.6M dataset.
        Bold scores indicate the best performing method for each metric and, where appropriate, scores whose means lie in the confidence interval of the best performing method.
    }
    \sisetup{detect-weight, table-align-uncertainty=true, mode=text}
    \renewrobustcmd{\bfseries}{\fontseries{b}\selectfont}
    \renewrobustcmd{\boldmath}{}
    \centering
    \vspace{0.1in}
    \begin{tabular}{lS[table-format=2.2(2)]S[table-format=1.4(2)]S[table-format=1.4(2)]}
        \toprule
        Models & {PSNR} & {SSIM} & {LPIPS} \tabularnewline
        \midrule
        StructVRNN & 24.46 \pm 0.17 & 0.8868 \pm 0.0025 & 0.0557 \pm 0.0013 \tabularnewline
        Ours & \bfseries 25.30 \pm 0.19 & \bfseries 0.9074 \pm 0.0022 & \bfseries 0.0509 \pm 0.0013 \tabularnewline
        Ours - $\frac{\Delta t}{2}$ & \bfseries 25.14 \pm 0.21 & 0.9049 \pm 0.0024 & 0.0534 \pm 0.0015 \tabularnewline
        Ours - MLP & 25.00 \pm 0.19 & 0.9047 \pm 0.0021 & 0.0529 \pm 0.0013 \tabularnewline
        Ours - GRU & 23.54 \pm 0.18 & 0.8868 \pm 0.0022 & 0.0683 \pm 0.0014 \tabularnewline
        \bottomrule
    \end{tabular}
\end{table}

\begin{table}
    \caption{
        \label{tab:res-bair}
        Numerical results (mean and $95\%$-confidence interval, when relevant) with respect to PSNR, SSIM, and LPIPS of tested methods on the BAIR dataset.
        Bold scores indicate the best performing method for each metric and, where appropriate, scores whose means lie in the confidence interval of the best performing method.
    }
    \sisetup{detect-weight, table-align-uncertainty=true, mode=text}
    \renewrobustcmd{\bfseries}{\fontseries{b}\selectfont}
    \renewrobustcmd{\boldmath}{}
    \centering
    \vspace{0.1in}
    \begin{tabular}{lS[table-format=2.2(2)]S[table-format=1.4(2)]S[table-format=1.4(2)]}
        \toprule
        Models & {PSNR} & {SSIM} & {LPIPS} \tabularnewline
        \midrule
        SV2P & \bfseries 20.39 \pm 0.27 & \bfseries 0.8169 \pm 0.0086 & 0.0912 \pm 0.0053 \tabularnewline
        SAVP & 18.44 \pm 0.25 & 0.7887 \pm 0.0092 & 0.0634 \pm 0.0026 \tabularnewline
        SVG & 18.95 \pm 0.26 & 0.8058 \pm 0.0088 & 0.0609 \pm 0.0034 \tabularnewline
        Ours & 19.59 \pm 0.27 & \bfseries 0.8196 \pm 0.0084 & \bfseries 0.0574 \pm 0.0032 \tabularnewline
        Ours - $\frac{\Delta t}{2}$ & 19.45 \pm 0.26 & \bfseries 0.8196 \pm 0.0082 & \bfseries 0.0579 \pm 0.0032 \tabularnewline
        Ours - MLP & 19.56 \pm 0.26 & \bfseries 0.8194 \pm 0.0084 & \bfseries 0.0572 \pm 0.0032 \tabularnewline
        Ours - GRU & 19.41 \pm 0.26 & \bfseries 0.8170 \pm 0.0084 & \bfseries 0.0585 \pm 0.0032 \tabularnewline
        \bottomrule
    \end{tabular}
\end{table}

\cref{tab:res-smmnist,tab:res-kth,tab:res-human,tab:res-bair} present, respectively, numerical results for PSNR, SSIM and LPIPS averaged over all time steps for our methods and considered baselines on the Moving MNIST, KTH, Human3.6M, and BAIR datasets, corresponding to \cref{fig:res-mmnist-2d,fig:res-kth-human-bair}.

Note that we choose the learned prior version of SVG for all datasets but KTH, for which we choose the fixed prior version, as done by its authors \citep{Denton2018}.

\section{Pendulum Experiments}
\label{app:Pendulum}

We test the ability of our model to model the dynamics of a common dataset used in the literature of state-space models \citep{Karl2017, Fraccaro2017}, Pendulum \citep{Karl2017}.
It consists of noisy observations of a dynamic torque-controlled pendulum; it is stochastic as the information of this control is not available.
We test our model, without the content variable $\bs{w}$, in the same setting as DVBF \citep{Karl2017} and KVAE \citep{Fraccaro2017} and report the corresponding ELBO scores in \cref{tab:res-pendulum}. The encoders and decoders for all methods are MLPs.

Our model outperforms DVBF and is merely beaten by KVAE.
This can be explained by the nature of the KVAE model, whose sequential model is learned using a Kalman filter rather than a VAE, allowing exact inference in the latent space.
On the contrary, DVBF is learned, like our model, by a sequential VAE, and is thus closer to our model than KVAE.
This result then shows that the dynamic model that we propose in the context of sequential VAEs is more adapted on this dataset than the one of DVBF, and achieve results close to a method taking advantage of exact inference using adapted tools such as Kalman filters.

\begin{table}
    \caption{
        \label{tab:res-pendulum}
        ELBO, in nats, achieved by DVBF, KVAE and our model on the Pendulum dataset.
        The bold score indicates the best performing method.
    }
    \sisetup{detect-weight, table-align-uncertainty=true, mode=text}
    \renewrobustcmd{\bfseries}{\fontseries{b}\selectfont}
    \renewrobustcmd{\boldmath}{}
    \vspace{0.1in}
    \centering
    \begin{tabular}{S[table-format=3.2]S[table-format=3.2]S[table-format=3.2]}
        \toprule
        {DVBF} & {KVAE} & {Ours} \tabularnewline
        \midrule
        \num{798.56} & \bfseries \num{807.02} & \num{806.12} \tabularnewline
        \bottomrule
    \end{tabular}
\end{table}

\section{Influence of the Euler step size}
\label{app:Oversampling}

\cref{tab:res-bair-os} details the numerical results of our model trained on BAIR with $\Delta t = \frac{1}{2}$ and tested with different values of $\Delta t$.
It shows that, when refining the Euler approximation, our model maintains its performances in settings unseen during training.

\cref{tab:res-kth-os-1,tab:res-kth-os-2} detail the numerical results of our model trained on KTH with, respectively, $\Delta t = 1$ and $\Delta t = \frac{1}{2}$, and tested with different values of $\Delta t$.
They show that if $\Delta t$ is chosen too high when training (here, $\Delta t = 1$), the model performance drops when refining the Euler approximation.
We assume that this phenomenon arises because the Euler approximation used in training is too rough, making the model adapt to an overly discretized dynamic that cannot be transferred to smaller Euler step sizes.
Indeed, when training with a smaller step size (here, $\Delta t = \frac{1}{2}$), results in the training settings are equivalent while results obtained with a lower $\Delta t$ are now much closer, if not equivalent, to the nominal ones.
This shows that the model learns a continuous dynamic if learned with a small enough step size.

Note that the loss of performance using a higher $\Delta t$ in testing than in training, like in \cref{tab:res-bair-os,tab:res-kth-os-2}, is expected as it corresponds to loosening the Euler approximation compared to training.
However, even in this challenging setting, our model maintains state-of-the-art results, demonstrating the quality of the learned dynamic as it can be further discretized if needed at the cost of a reasonable drop in performance.

\begin{table}
    \caption{
        \label{tab:res-bair-os}
        Numerical results for PSNR, SSIM, and LPIPS on BAIR of our model trained with $\Delta t = \frac{1}{2}$ and tested with different values of $\Delta t$.
    }
    \sisetup{detect-weight, table-align-uncertainty=true, mode=text}
    \renewrobustcmd{\bfseries}{\fontseries{b}\selectfont}
    \renewrobustcmd{\boldmath}{}
    \centering
    \vspace{0.1in}
    \begin{tabular}{lS[table-format=2.2(2)]S[table-format=1.4(2)]S[table-format=1.4(2)]}
        \toprule
        Step size $\Delta t$ & {PSNR} & {SSIM} & {LPIPS} \tabularnewline
        \midrule
        $\Delta t = 1$ & 18.95 \pm 0.25 & 0.8139 \pm 0.0081 & 0.0640 \pm 0.0036 \tabularnewline
        $\bs{\Delta t =} \frac{\bs{1}}{\bs{2}}$ & 19.59 \pm 0.27 & 0.8196 \pm 0.0084 & 0.0574 \pm 0.0032 \tabularnewline
        $\Delta t = \frac{1}{3}$ & 19.49 \pm 0.25 & 0.8201 \pm 0.0082 & 0.0574 \pm 0.0032 \tabularnewline
        $\Delta t = \frac{1}{4}$ & 19.45 \pm 0.26 & 0.8196 \pm 0.0082 & 0.0579 \pm 0.0032 \tabularnewline
        $\Delta t = \frac{1}{5}$ & 19.46 \pm 0.26 & 0.8197 \pm 0.0082 & 0.0584 \pm 0.0032 \tabularnewline
        \bottomrule
    \end{tabular}
\end{table}

\begin{table}
    \caption{
        \label{tab:res-kth-os-1}
        Numerical results for PSNR, SSIM, and LPIPS on KTH of our model trained with $\Delta t = 1$ and tested with different values of $\Delta t$.
    }
    \sisetup{detect-weight, table-align-uncertainty=true, mode=text}
    \renewrobustcmd{\bfseries}{\fontseries{b}\selectfont}
    \renewrobustcmd{\boldmath}{}
    \centering
    \vspace{0.1in}
    \begin{tabular}{lS[table-format=2.2(2)]S[table-format=1.4(2)]S[table-format=1.4(2)]}
        \toprule
        Step size $\Delta t$ & {PSNR} & {SSIM} & {LPIPS} \tabularnewline
        \midrule
        $\bs{\Delta t = 1}$ & 29.77 \pm 0.33 & 0.8681 \pm 0.0046 & 0.0742 \pm 0.0029 \tabularnewline
        $\Delta t = \frac{1}{2}$ & 29.18 \pm 0.35 & 0.8539 \pm 0.0054 & 0.0882 \pm 0.0040 \tabularnewline
        $\Delta t = \frac{1}{3}$ & 29.05 \pm 0.36 & 0.8509 \pm 0.0056 & 0.0924 \pm 0.0043 \tabularnewline
        $\Delta t = \frac{1}{4}$ & 28.98 \pm 0.37 & 0.8496 \pm 0.0057 & 0.0939 \pm 0.0045 \tabularnewline
        $\Delta t = \frac{1}{5}$ & 28.95 \pm 0.37 & 0.8490 \pm 0.0058 & 0.0948 \pm 0.0045 \tabularnewline
        \bottomrule
    \end{tabular}
\end{table}

\begin{table}
    \caption{
        \label{tab:res-kth-os-2}
        Numerical results for PSNR, SSIM, and LPIPS on KTH of our model trained with $\Delta t = \frac{1}{2}$ and tested with different values of $\Delta t$.
    }
    \sisetup{detect-weight, table-align-uncertainty=true, mode=text}
    \renewrobustcmd{\bfseries}{\fontseries{b}\selectfont}
    \renewrobustcmd{\boldmath}{}
    \centering
    \vspace{0.1in}
    \begin{tabular}{lS[table-format=2.2(2)]S[table-format=1.4(2)]S[table-format=1.4(2)]}
        \toprule
        Step size $\Delta t$ & {PSNR} & {SSIM} & {LPIPS} \tabularnewline
        \midrule
        $\Delta t = 1$ & 28.80 \pm 0.25 & 0.8495 \pm 0.0053 & 0.0994 \pm 0.0044 \tabularnewline
        $\bs{\Delta t =} \frac{\bs{1}}{\bs{2}}$ & 29.69 \pm 0.32 & 0.8697 \pm 0.0046 & 0.0736 \pm 0.0029 \tabularnewline
        $\Delta t = \frac{1}{3}$ & 29.52 \pm 0.33 & 0.8656 \pm 0.0048 & 0.0777 \pm 0.0033 \tabularnewline
        $\Delta t = \frac{1}{4}$ & 29.43 \pm 0.33 & 0.8633 \pm 0.0049 & 0.0790 \pm 0.0034 \tabularnewline
        $\Delta t = \frac{1}{5}$ & 29.35 \pm 0.34 & 0.8615 \pm 0.0050 & 0.0811 \pm 0.0036 \tabularnewline
        \bottomrule
    \end{tabular}
\end{table}

\section{Autoregressivity and Impact of Encoder and Decoder Architecture}
\label{app:Autoregressivity}

\begin{figure}
    \centering
    \includegraphics[width=\textwidth]{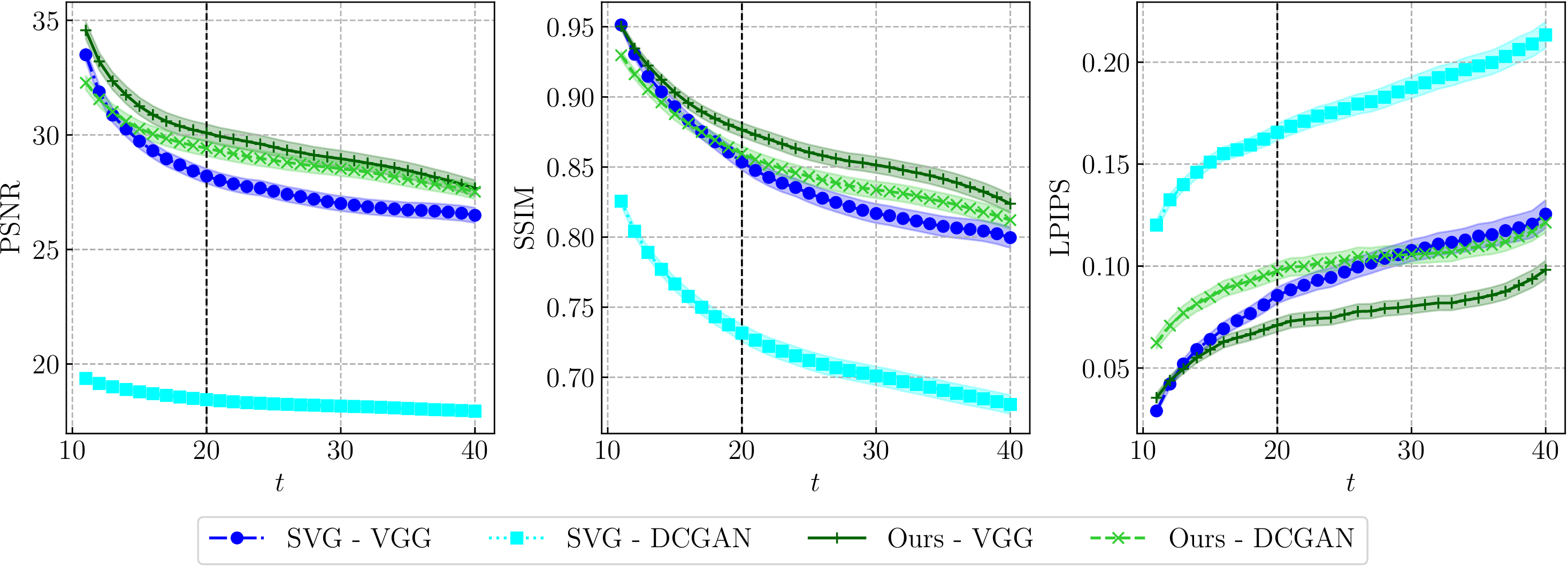}
    \vspace{-0.2in}
    \caption{
        \label{fig:res-kth-enc-pow}
        PNSR, SSIM and LPIPS scores with respect to $t$, with their $95\%$-confidence intervals, on the KTH dataset for SVG and our model with two choices of encoder and decoder architecture for each model: DCGAN and VGG.
        Vertical bars mark the length of training sequences.
    }
    \vspace{-0.13in}
\end{figure}

\cref{fig:res-kth-enc-pow,tab:fvd-dcgan} expose the numerical results on KTH of our model trained with $\Delta t = 1$ and SVG for two choices of encoder and decoder architectures: DCGAN and VGG.

Since DCGAN is a less powerful architecture than VGG, results of each method with VGG are expectedly better than those of the same method with DCGAN.
Moreover, our model outperforms SVG for any fixed choice of encoder and decoder architecture, which is coherent with \cref{fig:res-kth-human-bair}.

We observe, however, that the difference between a method using VGG and its DCGAN counterpart differs depending on the model.
Ours shows more robustness to the choice of encoder and decoder architecture, as it its performance decreases less than SVG when switching to a less powerful architecture.
This loss is particularly pronounced with respect to PSNR, which is the metric that most penalizes dynamics errors.
This shows that reducing the capacity of the encoders and decoders of SVG not only hurts its ability to produce realistic frames, as expected, but also substantially lowers its ability to learn a good dynamic.
We assume that this phenomenon is caused by the autoregressive nature of SVG, which makes it reliant on the performance of its encoders and decoders.
This supports our motivation to propose a non-autoregressive model for stochastic video prediction.

\begin{table*}
    \caption{
        \label{tab:fvd-dcgan}
        FVD scores for SVG and our method on KTH, trained either with DCGAN or VGG encoders and decoders, with their $95\%$-confidence intervals over five different samples from the models.
    }
    \sisetup{detect-weight, table-align-uncertainty=true, mode=text}
    \renewrobustcmd{\bfseries}{\fontseries{b}\selectfont}
    \renewrobustcmd{\boldmath}{}
    \centering
    \vspace{0.1in}
    \begin{tabular}{S[table-format=3(1)]S[table-format=3(1)]S[table-format=3(1)]S[table-format=3(1)]}
        \toprule
        \multicolumn{2}{c}{SVG} & \multicolumn{2}{c}{Ours} \tabularnewline
        \cmidrule(lr){1-2} \cmidrule(lr){3-4}
        {VGG} & {DCGAN} & {VGG} & {DCGAN} \tabularnewline
        \midrule
        377 \pm 6 & 542 \pm 6 & 220 \pm 2 & 371 \pm 3 \tabularnewline
        \bottomrule
    \end{tabular}
    \vspace{-0.1in}
\end{table*}

\section{Additional Samples}
\label{app:Samples}

This section includes additional samples corresponding to experiments described in \cref{sec:Experiments}.

\subsection{Stochastic Moving MNIST}

\begin{figure}
    \centering
    \scriptsize
    \begin{tabular}{rrl}
        \makecell{\sMnistTwoImg{cond}{1757}{0.204}} & \rotatebox[origin=c]{90}{\parbox[c]{0.8cm}{\centering Ground\\ Truth}} & \makecell{\sMnistTwoImg{ref_gt}{1757}{0.675}} \\
         & \rotatebox[origin=c]{90}{SVG} & \makecell{\sMnistTwoImg{ref_svg}{1757}{0.675}} \\
         & \rotatebox[origin=c]{90}{\parbox[c]{0.8cm}{\centering Ours\\ (Best)}} & \makecell{\sMnistTwoImg{hyp_best}{1757}{0.675}} \\
         & \rotatebox[origin=c]{90}{\parbox[c]{0.8cm}{\centering Ours\\ (Worst)}} & \makecell{\sMnistTwoImg{hyp_worst}{1757}{0.675}} \\
         & \rotatebox[origin=c]{90}{\parbox[c]{1cm}{\centering Ours\\ (Random)}} & \makecell{\sMnistTwoImg{hyp_random}{1757}{0.675}} \\
    \end{tabular}
    \caption{
        \label{fig:mmnist-2d-s-sample-2}
        Conditioning frames and corresponding ground truth and best samples with respect to PSNR from SVG and our method, and worst and random samples from our method, for an example of the Stochastic Moving MNIST dataset.
    }
\end{figure}

\begin{figure}
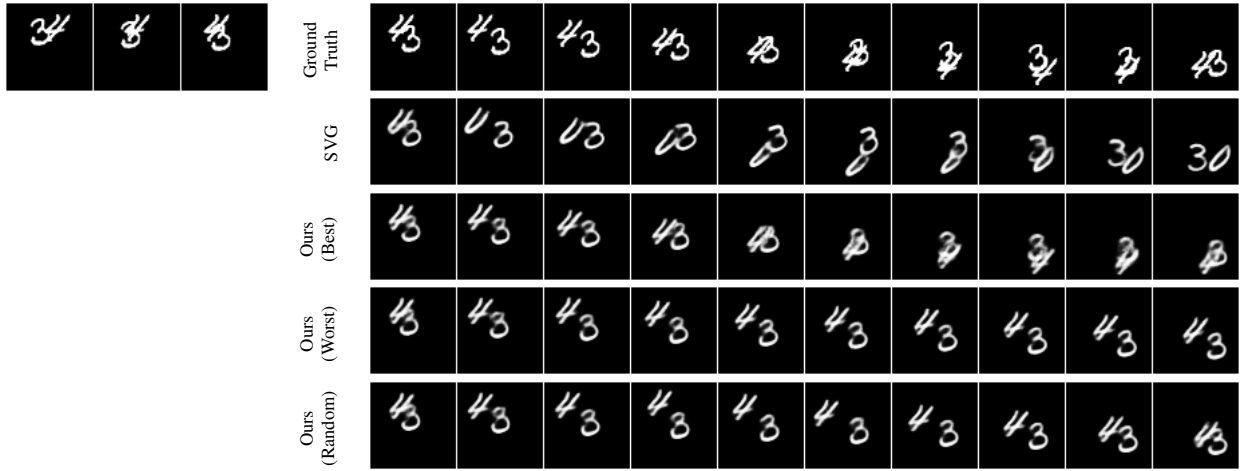

    \centering
    \scriptsize
    \begin{tabular}{rrl}
        \makecell{\sMnistTwoImg{cond}{1405}{0.204}} & \rotatebox[origin=c]{90}{\parbox[c]{0.8cm}{\centering Ground\\ Truth}} & \makecell{\sMnistTwoImg{ref_gt}{1405}{0.675}} \\
         & \rotatebox[origin=c]{90}{SVG} & \makecell{\sMnistTwoImg{ref_svg}{1405}{0.675}} \\
         & \rotatebox[origin=c]{90}{\parbox[c]{0.8cm}{\centering Ours\\ (Best)}} & \makecell{\sMnistTwoImg{hyp_best}{1405}{0.675}} \\
         & \rotatebox[origin=c]{90}{\parbox[c]{0.8cm}{\centering Ours\\ (Worst)}} & \makecell{\sMnistTwoImg{hyp_worst}{1405}{0.675}} \\
         & \rotatebox[origin=c]{90}{\parbox[c]{1cm}{\centering Ours\\ (Random)}} & \makecell{\sMnistTwoImg{hyp_random}{1405}{0.675}} \\
    \end{tabular}
    \caption{
        \label{fig:mmnist-2d-s-sample-3}
        Additional samples for the Stochastic Moving MNIST dataset (cf. \cref{fig:mmnist-2d-s-sample-2}).
    }
\end{figure}

\begin{figure}
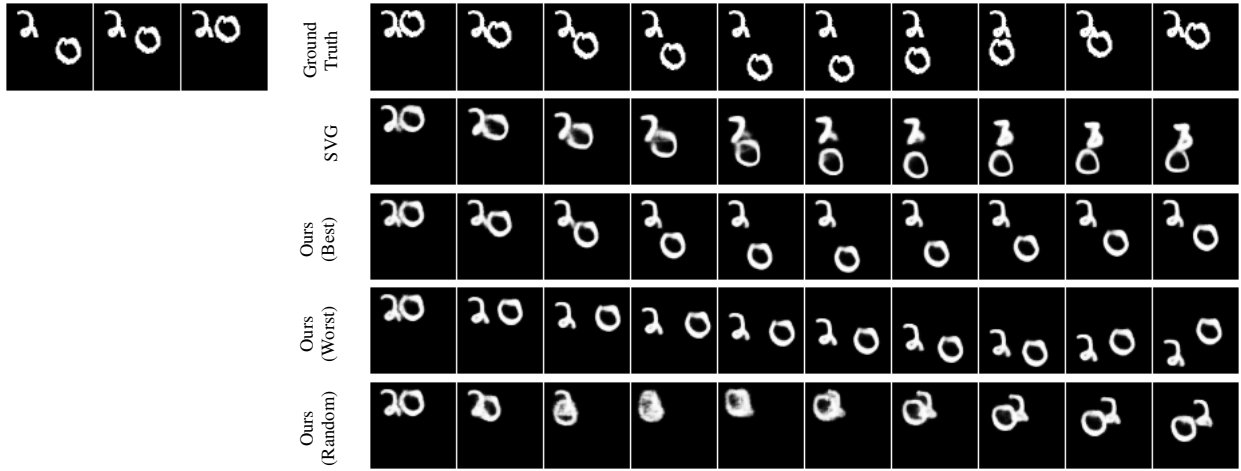

    \centering
    \scriptsize
    \begin{tabular}{rrl}
        \makecell{\sMnistTwoImg{cond}{4474}{0.204}} & \rotatebox[origin=c]{90}{\parbox[c]{0.8cm}{\centering Ground\\ Truth}} & \makecell{\sMnistTwoImg{ref_gt}{4474}{0.675}} \\
         & \rotatebox[origin=c]{90}{SVG} & \makecell{\sMnistTwoImg{ref_svg}{4474}{0.675}} \\
         & \rotatebox[origin=c]{90}{\parbox[c]{0.8cm}{\centering Ours\\ (Best)}} & \makecell{\sMnistTwoImg{hyp_best}{4474}{0.675}} \\
         & \rotatebox[origin=c]{90}{\parbox[c]{0.8cm}{\centering Ours\\ (Worst)}} & \makecell{\sMnistTwoImg{hyp_worst}{4474}{0.675}} \\
         & \rotatebox[origin=c]{90}{\parbox[c]{1cm}{\centering Ours\\ (Random)}} & \makecell{\sMnistTwoImg{hyp_random}{4474}{0.675}} \\
    \end{tabular}
    \caption{
        \label{fig:mmnist-2d-s-sample-changing-digit}
        Additional samples for the Stochastic Moving MNIST dataset (cf. \cref{fig:mmnist-2d-s-sample-2}).
        SVG fails to maintain the shape of a digit, while ours is temporally coherent.
    }
\end{figure}

\begin{figure}
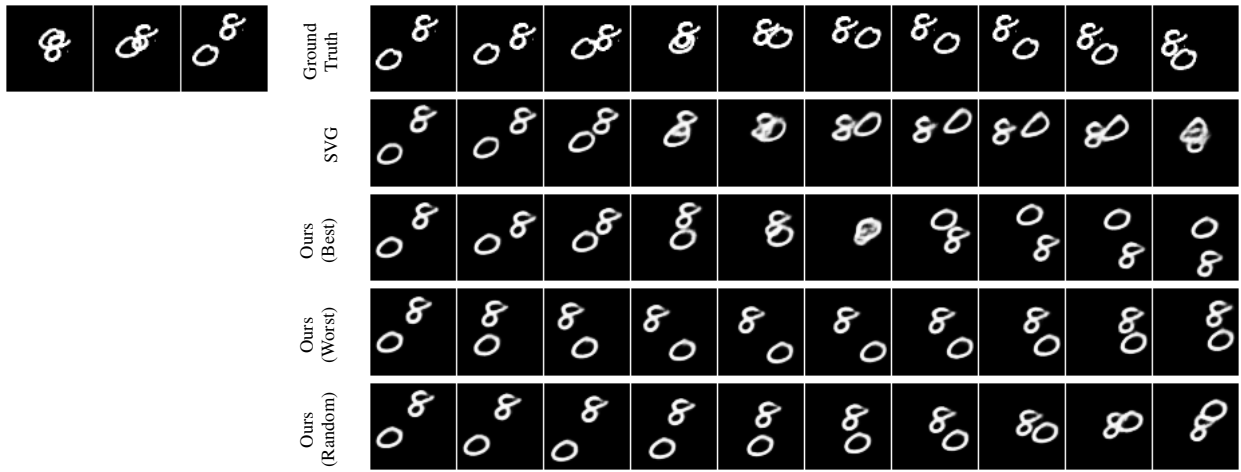

    \centering
    \scriptsize
    \begin{tabular}{rrl}
        \makecell{\sMnistTwoImg{cond}{4319}{0.204}} & \rotatebox[origin=c]{90}{\parbox[c]{0.8cm}{\centering Ground\\ Truth}} & \makecell{\sMnistTwoImg{ref_gt}{4319}{0.675}} \\
         & \rotatebox[origin=c]{90}{SVG} & \makecell{\sMnistTwoImg{ref_svg}{4319}{0.675}} \\
         & \rotatebox[origin=c]{90}{\parbox[c]{0.8cm}{\centering Ours\\ (Best)}} & \makecell{\sMnistTwoImg{hyp_best}{4319}{0.675}} \\
         & \rotatebox[origin=c]{90}{\parbox[c]{0.8cm}{\centering Ours\\ (Worst)}} & \makecell{\sMnistTwoImg{hyp_worst}{4319}{0.675}} \\
         & \rotatebox[origin=c]{90}{\parbox[c]{1cm}{\centering Ours\\ (Random)}} & \makecell{\sMnistTwoImg{hyp_random}{4319}{0.675}} \\
    \end{tabular}
    \caption{
        \label{fig:mmnist-2d-s-sample-worst-result}
        Additional samples for the Stochastic Moving MNIST dataset (cf. \cref{fig:mmnist-2d-s-sample-2}).
        This example was chosen in the worst $1\%$ testing examples of our model with respect to PSNR.
        Despite this adversarial criterion, our model maintains temporal consistency as digits are not deformed in the course of the video.
    }
\end{figure}

We present in \cref{fig:mmnist-2d-s-sample-2,fig:mmnist-2d-s-sample-3,fig:mmnist-2d-s-sample-changing-digit,fig:mmnist-2d-s-sample-worst-result} additional samples from SVG and our model on Stochastic Moving MNIST.

In particular, \cref{fig:mmnist-2d-s-sample-changing-digit} shows SVG changing a digit shape in the course of a prediction even though it does not cross another digit, whereas ours maintain the digit shape.
We assume that the advantage of our model comes from the latent nature of its dynamic and the use of a static content variable that is prevented from containing temporal information.
Indeed, even when the best sample from our model is not close from the ground truth of the dataset, like in \cref{fig:mmnist-2d-s-sample-worst-result}, it still maintains the shapes of the digits.

\subsection{KTH}

\begin{figure}
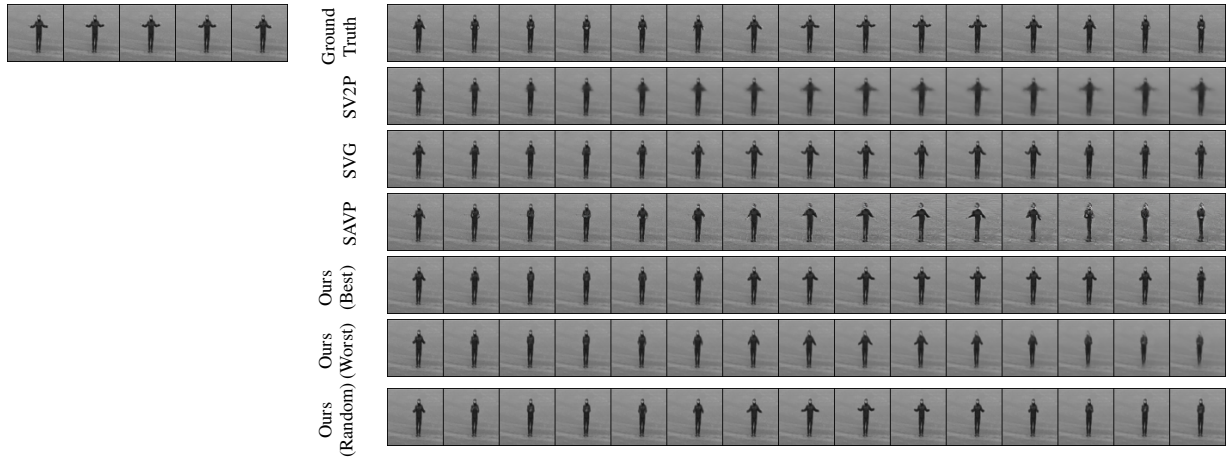

    \centering
    \scriptsize
    \begin{tabular}{rrl}
        \makecell{\kthImg{cond}{278}{0.217}} & \rotatebox[origin=c]{90}{\parbox[c]{0.8cm}{\centering Ground\\ Truth}} & \makecell{\kthImg{ref_gt}{278}{0.65}} \\
        & \rotatebox[origin=c]{90}{SV2P} & \makecell{\kthImg{ref_sv2p}{278}{0.65}} \\
        & \rotatebox[origin=c]{90}{SVG} & \makecell{\kthImg{ref_svg}{278}{0.65}} \\
        & \rotatebox[origin=c]{90}{SAVP} & \makecell{\kthImg{ref_savp}{278}{0.65}} \\
        & \rotatebox[origin=c]{90}{\parbox[c]{0.8cm}{\centering Ours\\ (Best)}} & \makecell{\kthImg{hyp_best}{278}{0.65}} \\
        & \rotatebox[origin=c]{90}{\parbox[c]{0.8cm}{\centering Ours\\ (Worst)}} & \makecell{\kthImg{hyp_worst}{278}{0.65}} \\
        & \rotatebox[origin=c]{90}{\parbox[c]{1cm}{\centering Ours\\ (Random)}} & \makecell{\kthImg{hyp_random}{278}{0.65}} \\
    \end{tabular}
    \caption{
        \label{fig:kth-sample-clapping}
        Conditioning frames and corresponding ground truth, best samples from SVG, SAVP and our method, and worst and random samples from our method, for an example of the KTH dataset.
        Samples are chosen according to their LPIPS with respect to the ground truth.
        On this specific task (clapping), all methods but SV2P (which produces blurry predictions) perform well, even though ours stays closer to the ground truth.
    }
\end{figure}

\begin{figure}
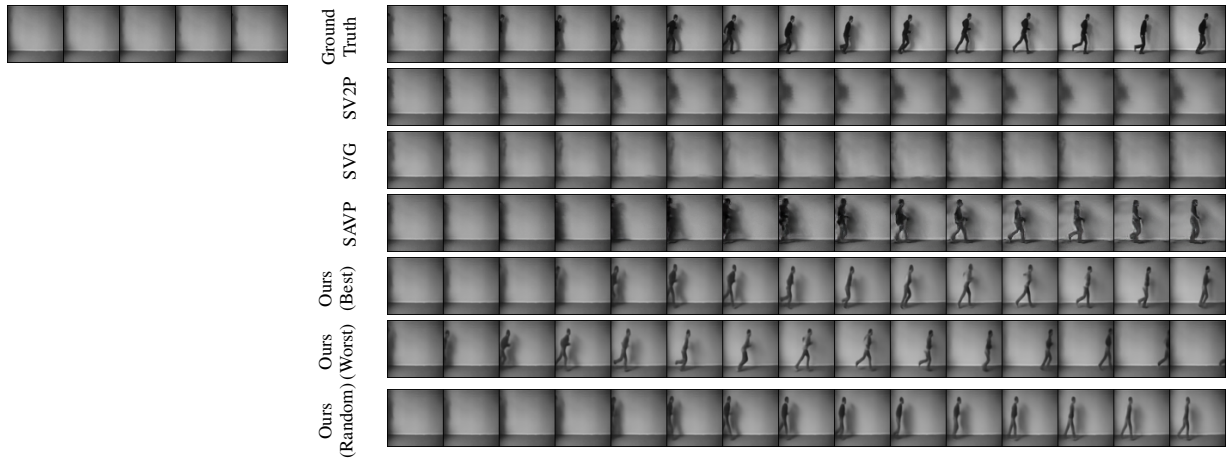

    \centering
    \scriptsize
    \begin{tabular}{rrl}
        \makecell{\kthImg{cond}{3}{0.217}} & \rotatebox[origin=c]{90}{\parbox[c]{0.8cm}{\centering Ground\\ Truth}} & \makecell{\kthImg{ref_gt}{3}{0.65}} \\
        & \rotatebox[origin=c]{90}{SV2P} & \makecell{\kthImg{ref_sv2p}{3}{0.65}} \\
        & \rotatebox[origin=c]{90}{SVG} & \makecell{\kthImg{ref_svg}{3}{0.65}} \\
        & \rotatebox[origin=c]{90}{SAVP} & \makecell{\kthImg{ref_savp}{3}{0.65}} \\
        & \rotatebox[origin=c]{90}{\parbox[c]{0.8cm}{\centering Ours\\ (Best)}} & \makecell{\kthImg{hyp_best}{3}{0.65}} \\
        & \rotatebox[origin=c]{90}{\parbox[c]{0.8cm}{\centering Ours\\ (Worst)}} & \makecell{\kthImg{hyp_worst}{3}{0.65}} \\
        & \rotatebox[origin=c]{90}{\parbox[c]{1cm}{\centering Ours\\ (Random)}} & \makecell{\kthImg{hyp_random}{3}{0.65}} \\
    \end{tabular}
    \caption{
        \label{fig:kth-sample-shadow}
        Additional samples for the KTH dataset (cf. \cref{fig:kth-sample-clapping}).
        In this example, the shadow of the subject is visible in the last conditioning frames, foreshadowing its appearance.
        This is a failure case for SV2P and SVG which only produce an indistinct shadow, whereas SAVP and our model make the subject appear.
        Yet, SAVP produces the wrong action and an inconsistent subject in its best sample, while ours is correct.
    }
\end{figure}

\begin{figure}
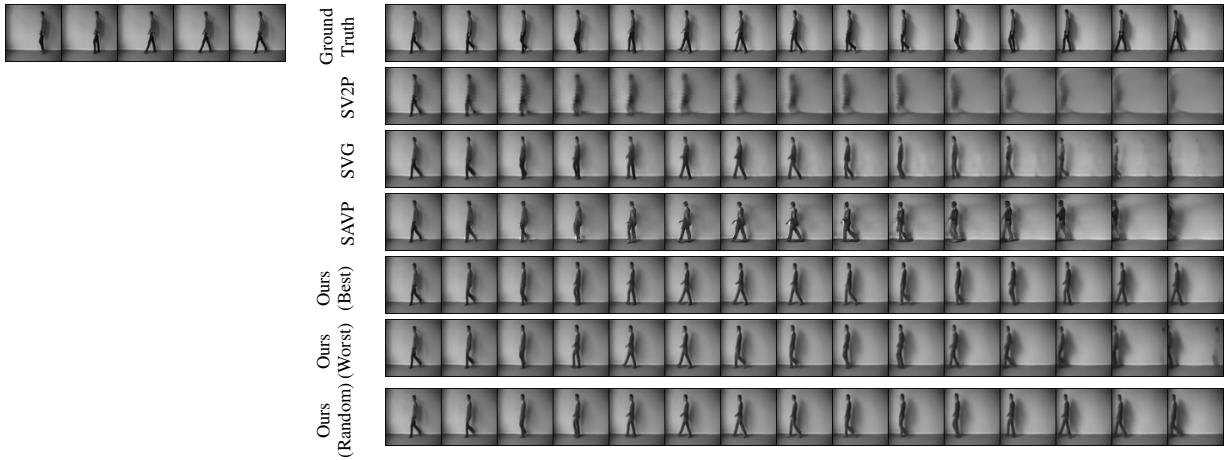

    \centering
    \scriptsize
    \begin{tabular}{rrl}
        \makecell{\kthImg{cond}{56}{0.217}} & \rotatebox[origin=c]{90}{\parbox[c]{0.8cm}{\centering Ground\\ Truth}} & \makecell{\kthImg{ref_gt}{56}{0.65}} \\
        & \rotatebox[origin=c]{90}{SV2P} & \makecell{\kthImg{ref_sv2p}{56}{0.65}} \\
        & \rotatebox[origin=c]{90}{SVG} & \makecell{\kthImg{ref_svg}{56}{0.65}} \\
        & \rotatebox[origin=c]{90}{SAVP} & \makecell{\kthImg{ref_savp}{56}{0.65}} \\
        & \rotatebox[origin=c]{90}{\parbox[c]{0.8cm}{\centering Ours\\ (Best)}} & \makecell{\kthImg{hyp_best}{56}{0.65}} \\
        & \rotatebox[origin=c]{90}{\parbox[c]{0.8cm}{\centering Ours\\ (Worst)}} & \makecell{\kthImg{hyp_worst}{56}{0.65}} \\
        & \rotatebox[origin=c]{90}{\parbox[c]{1cm}{\centering Ours\\ (Random)}} & \makecell{\kthImg{hyp_random}{56}{0.65}} \\
    \end{tabular}
    \caption{
        \label{fig:kth-sample-walking}
        Additional samples for the KTH dataset (cf. \cref{fig:kth-sample-clapping}).
        This example is a failure case for all methods: SV2P produces blurry frames, SVG and SAVP are not consistent (change of action or subject appearance in the video), and our model produces a ghost image at the end of the prediction on the worst sample only.
    }
\end{figure}

\begin{figure}
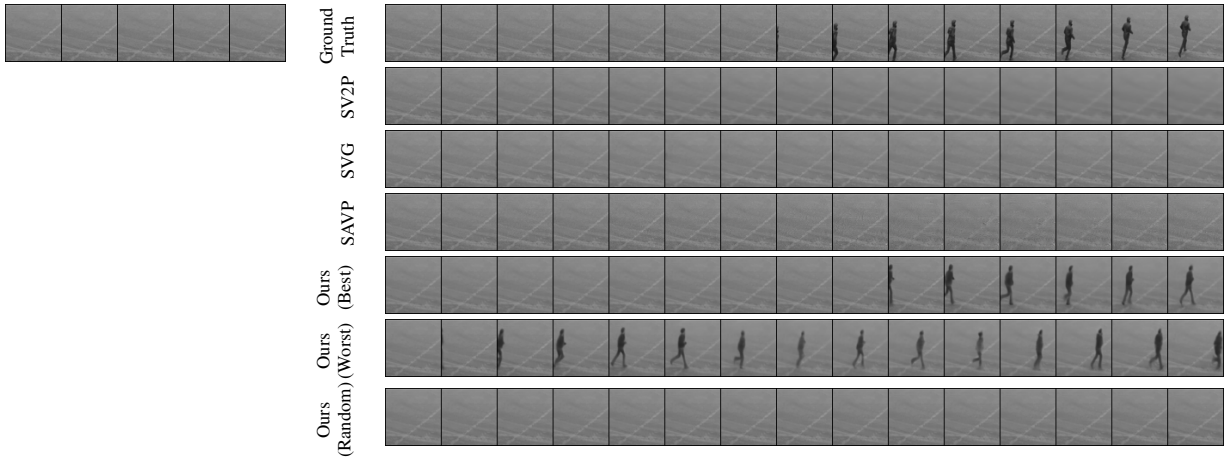

    \centering
    \scriptsize
    \begin{tabular}{rrl}
        \makecell{\kthImg{cond}{798}{0.217}} & \rotatebox[origin=c]{90}{\parbox[c]{0.8cm}{\centering Ground\\ Truth}} & \makecell{\kthImg{ref_gt}{798}{0.65}} \\
        & \rotatebox[origin=c]{90}{SV2P} & \makecell{\kthImg{ref_sv2p}{798}{0.65}} \\
        & \rotatebox[origin=c]{90}{SVG} & \makecell{\kthImg{ref_svg}{798}{0.65}} \\
        & \rotatebox[origin=c]{90}{SAVP} & \makecell{\kthImg{ref_savp}{798}{0.65}} \\
        & \rotatebox[origin=c]{90}{\parbox[c]{0.8cm}{\centering Ours\\ (Best)}} & \makecell{\kthImg{hyp_best}{798}{0.65}} \\
        & \rotatebox[origin=c]{90}{\parbox[c]{0.8cm}{\centering Ours\\ (Worst)}} & \makecell{\kthImg{hyp_worst}{798}{0.65}} \\
        & \rotatebox[origin=c]{90}{\parbox[c]{1cm}{\centering Ours\\ (Random)}} & \makecell{\kthImg{hyp_random}{798}{0.65}} \\
    \end{tabular}
    \caption{
        \label{fig:kth-sample-running}
        Additional samples for the KTH dataset (cf. \cref{fig:kth-sample-clapping}).
        Our model is the only one to make a subject appear, like in the ground truth.
    }
\end{figure}

\begin{figure}
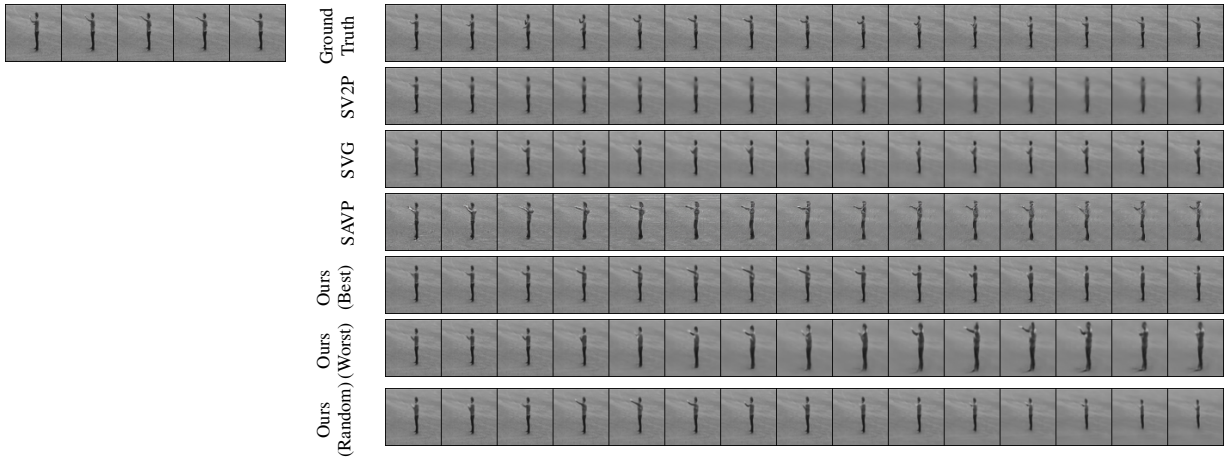

    \centering
    \scriptsize
    \begin{tabular}{rrl}
        \makecell{\kthImg{cond}{911}{0.217}} & \rotatebox[origin=c]{90}{\parbox[c]{0.8cm}{\centering Ground\\ Truth}} & \makecell{\kthImg{ref_gt}{911}{0.65}} \\
        & \rotatebox[origin=c]{90}{SV2P} & \makecell{\kthImg{ref_sv2p}{911}{0.65}} \\
        & \rotatebox[origin=c]{90}{SVG} & \makecell{\kthImg{ref_svg}{911}{0.65}} \\
        & \rotatebox[origin=c]{90}{SAVP} & \makecell{\kthImg{ref_savp}{911}{0.65}} \\
        & \rotatebox[origin=c]{90}{\parbox[c]{0.8cm}{\centering Ours\\ (Best)}} & \makecell{\kthImg{hyp_best}{911}{0.65}} \\
        & \rotatebox[origin=c]{90}{\parbox[c]{0.8cm}{\centering Ours\\ (Worst)}} & \makecell{\kthImg{hyp_worst}{911}{0.65}} \\
        & \rotatebox[origin=c]{90}{\parbox[c]{1cm}{\centering Ours\\ (Random)}} & \makecell{\kthImg{hyp_random}{911}{0.65}} \\
    \end{tabular}
    \caption{
        \label{fig:kth-sample-boxing}
        Additional samples for the KTH dataset (cf. \cref{fig:kth-sample-clapping}).
        The subject in this example is boxing, which is a challenging action in the dataset as all methods are far from the ground truth, even though ours remain closer in this case as well.
    }
\end{figure}

We present in \cref{fig:kth-sample-clapping,fig:kth-sample-shadow,fig:kth-sample-walking,fig:kth-sample-running,fig:kth-sample-boxing} additional samples from SV2P, SVG, SAVP and our model on KTH, with additional insights.

\subsection{Human3.6M}

\begin{figure}
    \centering
    \scriptsize
    \begin{tabular}{rrl}
        \makecell{\includegraphics[width=0.156\textwidth]{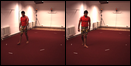}} & \rotatebox[origin=c]{90}{\parbox[c]{0.8cm}{\centering Ground\\ Truth}} & \makecell{\includegraphics[width=0.7\textwidth]{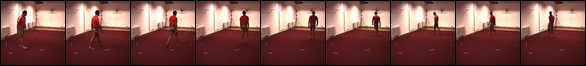}} \\
        & \rotatebox[origin=c]{90}{S-VRNN} & \makecell{\includegraphics[width=0.7\textwidth]{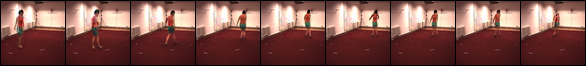}} \\
        & \rotatebox[origin=c]{90}{\parbox[c]{0.8cm}{\centering Ours\\ (Best)}} & \makecell{\includegraphics[width=0.7\textwidth]{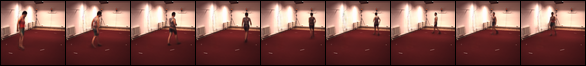}} \\
        & \rotatebox[origin=c]{90}{\parbox[c]{0.8cm}{\centering Ours\\ (Worst)}} & \makecell{\includegraphics[width=0.7\textwidth]{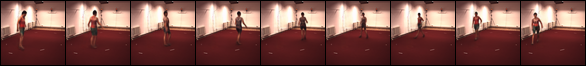}} \\
        & \rotatebox[origin=c]{90}{\parbox[c]{1cm}{\centering Ours\\ (Random)}} & \makecell{\includegraphics[width=0.7\textwidth]{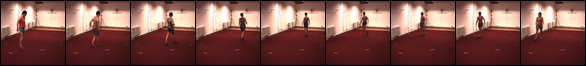}} \\
    \end{tabular}
    \caption{
        \label{fig:human-sample-2}
        Conditioning frames and corresponding ground truth, best samples from StructVRNN and our method, and worst and random samples from our method, for an example of the Human3.6M dataset.
        Samples are chosen according to their LPIPS with respect to the ground truth.
        We better capture the movements of the subject as well as their diversity, predict more realistic subjects, and present frames with less artefacts.
    }
\end{figure}

\begin{figure}
    \centering
    \scriptsize
    \begin{tabular}{rrl}
        \makecell{\includegraphics[width=0.156\textwidth]{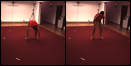}} & \rotatebox[origin=c]{90}{\parbox[c]{0.8cm}{\centering Ground\\ Truth}} & \makecell{\includegraphics[width=0.7\textwidth]{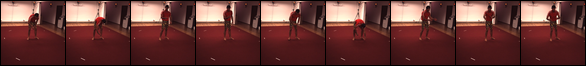}} \\
        & \rotatebox[origin=c]{90}{S-VRNN} & \makecell{\includegraphics[width=0.7\textwidth]{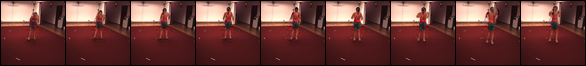}} \\
        & \rotatebox[origin=c]{90}{\parbox[c]{0.8cm}{\centering Ours\\ (Best)}} & \makecell{\includegraphics[width=0.7\textwidth]{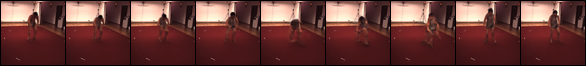}} \\
        & \rotatebox[origin=c]{90}{\parbox[c]{0.8cm}{\centering Ours\\ (Worst)}} & \makecell{\includegraphics[width=0.7\textwidth]{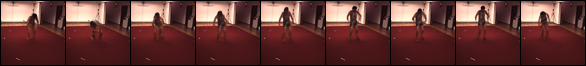}} \\
        & \rotatebox[origin=c]{90}{\parbox[c]{1cm}{\centering Ours\\ (Random)}} & \makecell{\includegraphics[width=0.7\textwidth]{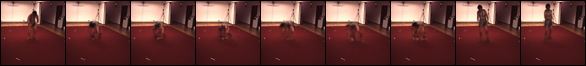}} \\
    \end{tabular}
    \caption{
        \label{fig:human-sample-3}
        Additional samples for the Human3.6M dataset (cf. \cref{fig:human-sample-2}).
        This action is better captured by our model, which is able to produce diverse realistic predictions.
    }
\end{figure}

We present in \cref{fig:human-sample-2,fig:human-sample-3} additional samples from StructVRNN and our model on Human3.6M, with additional insights.

\subsection{BAIR}

\begin{figure}
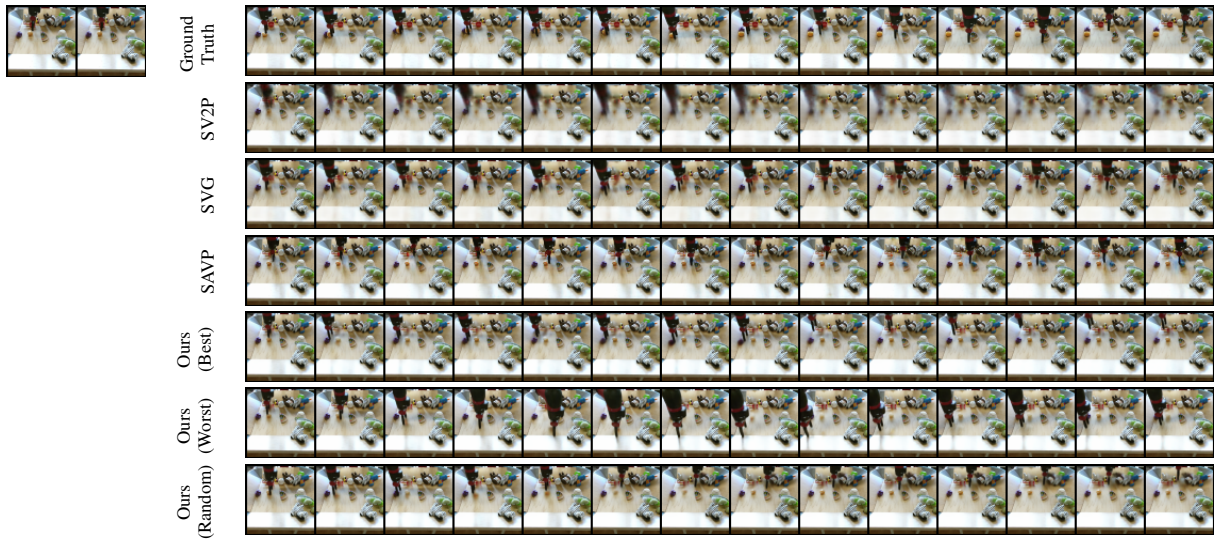

    \centering
    \scriptsize
    \begin{tabular}{rrl}
        \makecell{\bairImg{cond}{74}{0.108}} & \rotatebox[origin=c]{90}{\parbox[c]{0.8cm}{\centering Ground\\ Truth}} & \makecell{\bairImg{ref_gt}{74}{0.75}} \\
        & \rotatebox[origin=c]{90}{SV2P} & \makecell{\bairImg{ref_sv2p}{74}{0.75}} \\
        & \rotatebox[origin=c]{90}{SVG} & \makecell{\bairImg{ref_svg}{74}{0.75}} \\
        & \rotatebox[origin=c]{90}{SAVP} & \makecell{\bairImg{ref_savp}{74}{0.75}} \\
        & \rotatebox[origin=c]{90}{\parbox[c]{0.8cm}{\centering Ours\\ (Best)}} & \makecell{\bairImg{hyp_best}{74}{0.75}} \\
        & \rotatebox[origin=c]{90}{\parbox[c]{0.8cm}{\centering Ours\\ (Worst)}} & \makecell{\bairImg{hyp_worst}{74}{0.75}} \\
        & \rotatebox[origin=c]{90}{\parbox[c]{1cm}{\centering Ours\\ (Random)}} & \makecell{\bairImg{hyp_random}{74}{0.75}} \\
    \end{tabular}
    \caption{
        \label{fig:bair-sample-1}
        Conditioning frames and corresponding ground truth, best samples from SVG, SAVP and our method, and worst and random samples from our method, for an example of the BAIR dataset.
        Samples are chosen according to their LPIPS with respect to the ground truth.
    }
\end{figure}

\begin{figure}
    \centering
    \scriptsize
    \begin{tabular}{rrl}
        \makecell{\bairImg{cond}{155}{0.108}} & \rotatebox[origin=c]{90}{\parbox[c]{0.8cm}{\centering Ground\\ Truth}} & \makecell{\bairImg{ref_gt}{155}{0.75}} \\
        & \rotatebox[origin=c]{90}{SV2P} & \makecell{\bairImg{ref_sv2p}{155}{0.75}} \\
        & \rotatebox[origin=c]{90}{SVG} & \makecell{\bairImg{ref_svg}{155}{0.75}} \\
        & \rotatebox[origin=c]{90}{SAVP} & \makecell{\bairImg{ref_savp}{155}{0.75}} \\
        & \rotatebox[origin=c]{90}{\parbox[c]{0.8cm}{\centering Ours\\ (Best)}} & \makecell{\bairImg{hyp_best}{155}{0.75}} \\
        & \rotatebox[origin=c]{90}{\parbox[c]{0.8cm}{\centering Ours\\ (Worst)}} & \makecell{\bairImg{hyp_worst}{155}{0.75}} \\
        & \rotatebox[origin=c]{90}{\parbox[c]{1cm}{\centering Ours\\ (Random)}} & \makecell{\bairImg{hyp_random}{155}{0.75}} \\
    \end{tabular}
    \caption{
        \label{fig:bair-sample-2}
        Additional samples for the BAIR dataset (cf. \cref{fig:bair-sample-1}).
    }
\end{figure}

\begin{figure}
    \centering
    \scriptsize
    \begin{tabular}{rrl}
        \makecell{\bairImg{cond}{38}{0.108}} & \rotatebox[origin=c]{90}{\parbox[c]{0.8cm}{\centering Ground\\ Truth}} & \makecell{\bairImg{ref_gt}{38}{0.75}} \\
        & \rotatebox[origin=c]{90}{SV2P} & \makecell{\bairImg{ref_sv2p}{38}{0.75}} \\
        & \rotatebox[origin=c]{90}{SVG} & \makecell{\bairImg{ref_svg}{38}{0.75}} \\
        & \rotatebox[origin=c]{90}{SAVP} & \makecell{\bairImg{ref_savp}{38}{0.75}} \\
        & \rotatebox[origin=c]{90}{\parbox[c]{0.8cm}{\centering Ours\\ (Best)}} & \makecell{\bairImg{hyp_best}{38}{0.75}} \\
        & \rotatebox[origin=c]{90}{\parbox[c]{0.8cm}{\centering Ours\\ (Worst)}} & \makecell{\bairImg{hyp_worst}{38}{0.75}} \\
        & \rotatebox[origin=c]{90}{\parbox[c]{1cm}{\centering Ours\\ (Random)}} & \makecell{\bairImg{hyp_random}{38}{0.75}} \\
    \end{tabular}
    \caption{
        \label{fig:bair-sample-3}
        Additional samples for the BAIR dataset (cf. \cref{fig:bair-sample-1}).
    }
\end{figure}

We present in \cref{fig:bair-sample-1,fig:bair-sample-2,fig:bair-sample-3} additional samples from SV2P, SVG, SAVP and our model on BAIR.

\subsection{Oversampling}

\begin{figure}
    \centering
    \subfigure[Cropped KTH sample, centered on the subject.]{\kthImg{hyp_2_dt}{955}{0.6}}
    \subfigure[Cropped Human3.6M sample, centered on the subject.]{\includegraphics[width=0.6\textwidth]{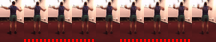}}
    \subfigure[Cropped BAIR sample, centered on the robot arm.]{\bairImg{hyp_2_dt}{5}{0.6}}
    
    \caption{
        \label{fig:oversampling-2}
        Generation examples at doubled or quadrupled frame rate, using a halved $\Delta t$ compared to training.
        Frames including a bottom red dashed bar are intermediate frames.
    }
\end{figure}

We present in \cref{fig:oversampling-2} additional examples of video generation at a doubled and quadrupled frame rate by our model.

\subsection{Content Swap}

\begin{figure}
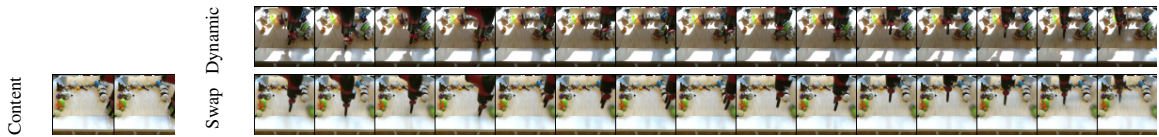

    \centering
    \scriptsize
    \begin{tabular}{rrrl}
        & & \rotatebox[origin=c]{90}{Dynamic} & \makecell{\bairImg{rec_pos}{136-143}{0.7}} \\
        \rotatebox[origin=c]{90}{Content} & \makecell{\bairImg{rec_con}{136-143}{0.094}} & \rotatebox[origin=c]{90}{Swap} & \makecell{\bairImg{rec_swap}{136-143}{0.7}}
    \end{tabular}
    \caption{
        \label{fig:bair-content-swap-1}
        Video (bottom right) generated from the combination of dynamic variables ($\bs{y}$, $\bs{z}$) inferred with a video (top) and a content variable ($\bs{w}$) computed with the conditioning frames of another video (bottom left).
    }
\end{figure}

\begin{figure}
    \centering
    \scriptsize
    \begin{tabular}{rrrl}
        & & \rotatebox[origin=c]{90}{Dynamic} & \makecell{\bairImg{rec_pos}{88-154}{0.7}} \\
        \rotatebox[origin=c]{90}{Content} & \makecell{\bairImg{rec_con}{88-154}{0.094}} & \rotatebox[origin=c]{90}{Swap} & \makecell{\bairImg{rec_swap}{88-154}{0.7}}
    \end{tabular}
    \caption{
        \label{fig:bair-content-swap-2}
        Additional example of content swap (cf. \cref{fig:bair-content-swap-1}).
    }
\end{figure}

\begin{figure}
    \centering
    \scriptsize
    \begin{tabular}{rrrl}
        & & \rotatebox[origin=c]{90}{Dynamic} &\makecell{\kthImg{rec_pos}{66-578}{0.7}} \\
        \rotatebox[origin=c]{90}{Content} &\makecell{\kthImg{rec_con}{66-578}{0.14}} & \rotatebox[origin=c]{90}{Swap} & \makecell{\kthImg{rec_swap}{66-578}{0.7}}
    \end{tabular}
    \caption{
        \label{fig:kth-content-swap-1}
        Additional example of content swap (cf. \cref{fig:bair-content-swap-1}).
        In this example, the extracted content is the video background, which is successfully transferred to the target video.
    }
\end{figure}

\begin{figure}
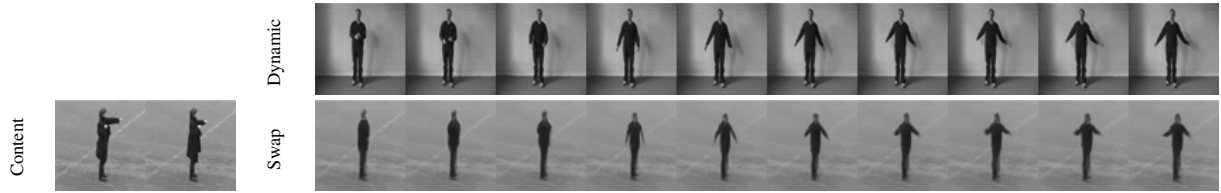

    \centering
    \scriptsize
    \begin{tabular}{rrrl}
        & & \rotatebox[origin=c]{90}{Dynamic} &\makecell{\kthImg{rec_pos}{836-187}{0.7}} \\
        \rotatebox[origin=c]{90}{Content} &\makecell{\kthImg{rec_con}{836-187}{0.14}} & \rotatebox[origin=c]{90}{Swap} & \makecell{\kthImg{rec_swap}{836-187}{0.7}}
    \end{tabular}
    \caption{
        \label{fig:kth-content-swap-2}
        Additional example of content swap (cf. \cref{fig:bair-content-swap-1}).
        In this example, the extracted content is the video background and the subject appearance, which are successfully transferred to the target video.
    }
\end{figure}

\begin{figure}
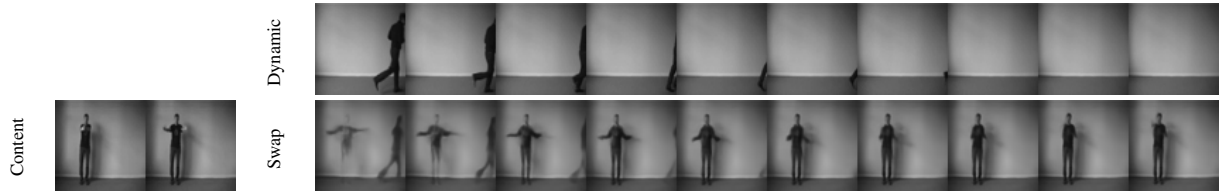

    \centering
    \scriptsize
    \begin{tabular}{rrrl}
        & & \rotatebox[origin=c]{90}{Dynamic} &\makecell{\kthImg{rec_pos}{96-79}{0.7}} \\
        \rotatebox[origin=c]{90}{Content} &\makecell{\kthImg{rec_con}{96-79}{0.14}} & \rotatebox[origin=c]{90}{Swap} & \makecell{\kthImg{rec_swap}{96-79}{0.7}}
    \end{tabular}
    \caption{
        \label{fig:kth-content-swap-fail}
        Additional example of content swap (cf. \cref{fig:bair-content-swap-1}).
        This example shows a failure case of content swapping.
    }
\end{figure}

\begin{figure}
    \centering
    \scriptsize
    \begin{tabular}{rrrl}
        & & \rotatebox[origin=c]{90}{Dynamic} & \makecell{\includegraphics[width=0.75\textwidth]{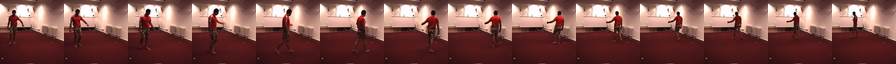}} \\
        \rotatebox[origin=c]{90}{Content} & \makecell{\includegraphics[width=0.107\textwidth]{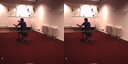}} & \rotatebox[origin=c]{90}{Swap} & \makecell{\includegraphics[width=0.75\textwidth]{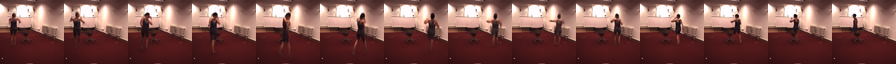}}
    \end{tabular}
    \caption{
        \label{fig:human-content-swap-2}
        Additional example of content swap (cf. \cref{fig:bair-content-swap-1}).
    }
\end{figure}

We present in \cref{fig:bair-content-swap-1,fig:bair-content-swap-2,fig:kth-content-swap-1,fig:kth-content-swap-2,fig:kth-content-swap-fail,fig:human-content-swap-2} additional examples of content swap as in \cref{fig:human-content-swap-1}.

\subsection{Interpolation in the Latent Space}

\begin{figure}
    \centering
    \subfigure[Ref 1]{\includegraphics[width=0.1\textwidth]{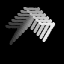}}
    \hfill
    \subfigure[Rec 1]{\includegraphics[width=0.1\textwidth]{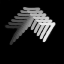}}
    \hfill
    \subfigure[Interpolation]{
        \includegraphics[width=0.1\textwidth]{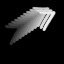}
        \includegraphics[width=0.1\textwidth]{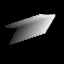}
        \includegraphics[width=0.1\textwidth]{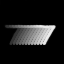}
        \includegraphics[width=0.1\textwidth]{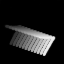}
        \includegraphics[width=0.1\textwidth]{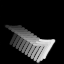}
    }
    \hfill
    \subfigure[Rec 2]{\includegraphics[width=0.1\textwidth]{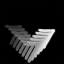}}
    \hfill
    \subfigure[Ref 2]{\includegraphics[width=0.1\textwidth]{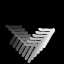}}
    \caption{
        \label{fig:mnist-interpolation-2}
        From left to right, $\bs{x}^{\mathrm{s}}$, $\widehat{\bs{x}}^{\mathrm{s}}$ (reconstruction of $\bs{x}^{\mathrm{s}}$ by the VAE of our model), results of the interpolation in the latent space between $\bs{x}^{\mathrm{s}}$ and $\bs{x}^{\mathrm{t}}$, $\widehat{\bs{x}}^{\mathrm{t}}$ and $\bs{x}^{\mathrm{t}}$.
        Each trajectory is materialized in shades of grey in the frames.
    }
\end{figure}

\begin{figure}
    \centering
    \subfigure[Ref 1]{\includegraphics[width=0.1\textwidth]{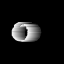}}
    \hfill
    \subfigure[Rec 1]{\includegraphics[width=0.1\textwidth]{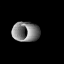}}
    \hfill
    \subfigure[Interpolation]{
        \includegraphics[width=0.1\textwidth]{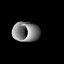}
        \includegraphics[width=0.1\textwidth]{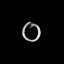}
        \includegraphics[width=0.1\textwidth]{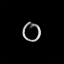}
        \includegraphics[width=0.1\textwidth]{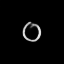}
        \includegraphics[width=0.1\textwidth]{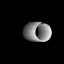}
    }
    \hfill
    \subfigure[Rec 2]{\includegraphics[width=0.1\textwidth]{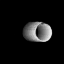}}
    \hfill
    \subfigure[Ref 2]{\includegraphics[width=0.1\textwidth]{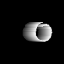}}
    \caption{
        \label{fig:mnist-interpolation-3}
        Additional example of interpolation in the latent space between two trajectories (cf. \cref{fig:mnist-interpolation-2}).
    }
\end{figure}

We present in \cref{fig:mnist-interpolation-2,fig:mnist-interpolation-3} additional examples of interpolation in the latent space between two trajectories as in \cref{fig:mnist-interpolation-1}.

\end{document}